\documentclass[conference]{IEEEtran}
\IEEEoverridecommandlockouts
\usepackage{cite}
\usepackage{amsmath,amssymb,amsfonts}
\usepackage{algorithmic}
\usepackage{graphicx}
\usepackage{textcomp}
\usepackage{xcolor}
\usepackage[hyphens]{url}

\usepackage{soul}

\makeatletter 
\newcount\SOUL@minus
\makeatother  

\IEEEpubid{\begin{minipage}{\textwidth}\ \\[1pt]
Accepted at the 49\textsuperscript{th} IEEE/ACM International Symposium on Computer \\
Architecture (ISCA '22) -- \href{https://doi.org/10.1145/3470496.3527438}{DOI: 10.1145/3470496.3527438 }
\end{minipage}} 

\usepackage{breqn}

\usepackage{subfig}
\usepackage{amsmath,amssymb,amsfonts}

\usepackage{amsmath}
\mathchardef\mhyphen="2D 
\usepackage{amsfonts}

\usepackage{xcolor}
\usepackage{listings}
\usepackage{graphicx}
\usepackage{amssymb}
\usepackage{pifont}
\newcommand{\cmark}{\ding{51}}%
\newcommand{\xmark}{\ding{55}}%
\usepackage{hyperref}
\usepackage[inline]{enumitem}

\usepackage{float}
\usepackage{textcomp}
\usepackage[linesnumbered, ruled]{algorithm2e}
\SetKwRepeat{Do}{do}{while}%
\usepackage{comment}
\usepackage{multicol}

\newsavebox{\tempboxa}
\newsavebox{\tempboxb}
\newsavebox{\tempboxc}
\usepackage{multirow}
\usepackage{booktabs}
\usepackage{xspace}

\usepackage{etoolbox}
\AtBeginEnvironment{align}{%
  \fontsize{9}{6}%
}

\newcommand{\ra}[1]{\renewcommand{\arraystretch}{#1}}

\definecolor{mGreen}{rgb}{0,0.6,0}
\definecolor{mGray}{rgb}{0.5,0.5,0.5}
\definecolor{mPurple}{rgb}{0.58,0,0.82}
\definecolor{backgroundColour}{rgb}{0.95,0.95,0.92}

\definecolor{fabulous}{rgb}{0.958, 0.188, 0.478}

\definecolor{BLU}{rgb}{0.01, 0.08, .9}

\newcommand*{\myfont}{\fontfamily{lmss}\selectfont}
\DeclareTextFontCommand{\HTfont}{\myfont}

\newcommand{\OURLCORE}{Mokey} 

\newcommand{\OURL}{\textit{\OURLCORE}\xspace} 

\def\BibTeX{{\rm B\kern-.05em{\sc i\kern-.025em b}\kern-.08em
    T\kern-.1667em\lower.7ex\hbox{E}\kern-.125emX}}

\begin{document}
\pagenumbering{arabic}
\title{Mokey: Enabling Narrow Fixed-Point Inference for Out-of-the-Box Floating-Point Transformer Models}
\author{\IEEEauthorblockN{Ali Hadi Zadeh}
\IEEEauthorblockA{
\textit{University of Toronto, Vector Institute}\\
Toronto, Canada \\
hadizade@ece.utoronto.ca}\\
\IEEEauthorblockN{Ameer Abdelhadi}
\IEEEauthorblockA{
\textit{University of Toronto}\\
Toronto, Canada \\
ameer.abdelhadi@utoronto.ca}
\and 
\IEEEauthorblockN{Mostafa Mahmoud}
\IEEEauthorblockA{
\textit{University of Toronto}\\
Toronto, Canada \\
mostafa.mahmoud@mail.utoronto.ca}\\
\IEEEauthorblockN{Andreas Moshovos}
\IEEEauthorblockA{
\textit{University of Toronto, Vector Institute}\\
Toronto, Canada \\
moshovos@ece.utoronto.ca}
}

\thispagestyle{plain}
\pagestyle{plain}

\maketitle

\IEEEpubidadjcol

\begin{abstract}
Increasingly larger and better Transformer models keep advancing state-of-the-art accuracy and capability for Natural Language Processing applications. These models demand more computational power, storage, and energy. \OURL reduces the footprint of state-of-the-art 32-bit or 16-bit floating-point transformer models by quantizing all values to 4-bit indexes into dictionaries of representative 16-bit fixed-point centroids.
\OURL does not need fine-tuning, an essential feature as often the training resources or datasets are not available to many.
Exploiting the range of values that naturally occur in transformer models, \OURL selects centroid values to also fit an exponential curve. This unique feature enables \OURL to replace the bulk of the original multiply-accumulate operations with narrow 3b fixed-point additions resulting in an area- and energy-efficient hardware accelerator design. Over a set of state-of-the-art transformer models, the \OURL accelerator delivers an order of magnitude improvements in energy efficiency over a Tensor Cores-based accelerator while improving performance by at least $4\times$ and as much as $15\times$ depending on the model and on-chip buffering capacity. Optionally, \OURL can be used as a memory compression assist for any other accelerator, transparently stashing wide floating-point or fixed-point activations or weights into narrow 4-bit indexes. \OURL proves superior to prior state-of-the-art quantization methods for Transformers. 
\end{abstract}

\section{Introduction}

Creating machines that can ``understand’’ our language and ``interact’’ with us as we interact with each other has always been a dream that motivated many and captured the imaginations of even more. Transformer models have demonstrated remarkable success towards this goal by tackling many natural language understanding tasks. Recently, Open-AI’s GPT-3~\cite{brown2020language}, with its roughly 175 billion parameters, and Google's Switch transformer with its 1.6 trillion parameters~\cite{switchTransformer}, proved that attention-based models --- given enough parameters --- can perform tasks that they have not been explicitly trained for. For example, with zero-shot learning~\cite{brown2020language,zeroshot1st}, a model that is pre-trained for language modeling on copious amounts of text (including Common Crawl's Petabyte datasets~\cite{commoncrawl}) could answer various questions, including solving a mathematical equation or writing an HTML script.

The development trend of these NLP models shows that the more the parameters, the more powerful the model becomes. Unfortunately, this comes at a cost: More parameters require more storage, data transfers, and computation~\cite{desislavov2021compute}. From 2018 with GPT~\cite{gpt} and BERT~\cite{BERT} with hundreds of millions of parameters to 2021 with the 1.6 trillion parameter Switch transformer, parameter footprint increased from hundreds of megabytes to terabytes. Computing hardware must meet their storage and compute needs and to do so within the constraints of each use case. As these models are massive and as the on-chip memory of modern silicon chips is limited, it is off-chip memory accesses that account for most of the execution time and energy consumption~\cite{ivanov2021data,GOBO,CompressingBERT}. 

\begin{figure}[t!]
\centering
\includegraphics[width=0.45\textwidth]{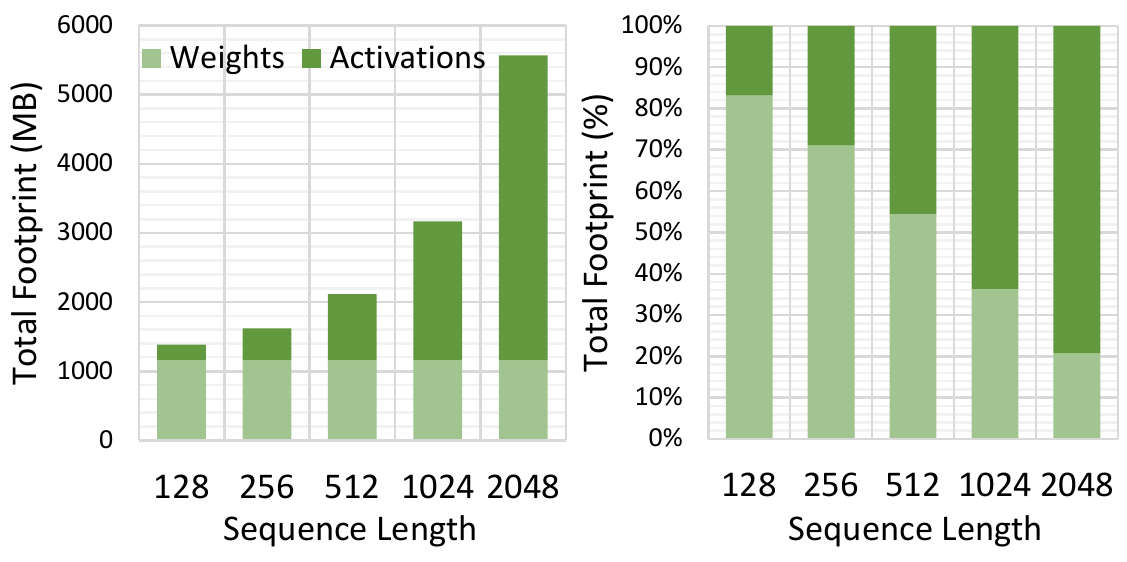}
\caption{BERT-Large: Weight and Activation Memory Footprint.} 
\label{fig:footprint_act}
\end{figure}

Transformer models also incur a quadratic growth in activation footprint when scaling the input sequence. For sequences of up to 128 tokens  (e.g., GLUE dataset tasks~\cite{glue}), buffering activations between layers requires anywhere between 1.5MB to 2MB depending on the model, layer, and dataflow. Keeping these activations on-chip is certainly possible and avoids the one to two orders of magnitude higher energy and latency costs of off-chip memory. However, when processing longer sequences such as multiple paragraphs or a book chapter, activation volume can easily exceed on-chip buffer capacities\cite{katharopoulos2020transformers,wang2020linformer,zaheer2021big,beltagy2020longformer}. As an example, Figure~\ref{fig:footprint_act} shows the total footprint of activations and weights as a function of sequence length for BERT-Large. When the sequence length exceeds 512 tokens, activations dominate total memory footprint.

Model compression reduces model size improving overall efficiency. Recent model compression works fall under three general classes: \textit{Pruning} which forces some weights or activations to zero~\cite{TPrune,kim2021learned,lagunas2021block,parnami2021pruning,li2021differentiable,peer2021greedy,shim2021layerwise,wang2021spatten} combined with ``zero-aware'' memory encoding. \textit{Knowledge distillation} distills a larger, ``teacher'' model into a smaller ``student'' model~\cite{sanh2019distilbert,aguilar2020knowledge,touvron2021training,masumura2021hierarchical,mukherjee2020distilling,liu2019mkd,jiang2021knowledge}. 
Lastly, \textit{quantization} where the parameters and/or activations are quantized to shorter bit-widths~\cite{GOBO,q-bert,intel8b,kim2021ibert,zhang2020ternarybert,chung2020extremely}. By default transformers use single-precision or 16b floating-point values.
Except for I-BERT~\cite{kim2021ibert} which quantizes all values to 8b fixed-point, quantization methods still require floating-point or mixed fixed-point and floating-point units. In addition, all aforementioned compression methods require fine-tuning --- further training for a few epochs over a previously trained model --- to compensate for lost accuracy. Fine-tuning is expensive, requires access to sufficient system resources~\cite{CostOfTrainingNLP2020} and worse to datasets that might be unavailable to end-users. Having to search for optimal hyperparameters and quantization per task further exacerbates the cost of fine-tuning. GOBO is a post-training compression method for weights only that requires no fine-tuning ~\cite{GOBO}. GOBO quantizes the vast majority of weights to 3b or 4b indexes into a dictionary of 8 or 16 representative floating-point values (centroids). The very few remaining outlier values remain as-is (in single-precision 32b floating-point/FP32 or FP16). Further, GOBO still requires floating-point units. Ultimately, a quantization method that 1)~does not require fine-tuning, 2)~quantizes weights and activations, and 3)~would require only fixed-point units, would improve overall energy efficiency and execution time performance. This is the goal of this work.

Quantizing activations presents additional challenges compared to quantizing weights only. First, since weights are statically known,  quantization is done \textit{off-line} using iterative centroid selection algorithms. The runtime and energy overheads of these algorithms are prohibitive for the \textit{runtime} calculated activations. Quantizing activations calls for light-weight, \textit{online} methods. Second, the value range of activations is much wider than those for weights (see Table~\ref{tbl:Relwork})~\cite{intel8b,q-bert,kim2021ibert,zhang2020ternarybert}. Most quantization methods still require 8 bits per activation or forego activation quantization altogether.  

This work introduces \OURL, a \textit{post-training} transformer model compression method for activations and weights that reduces memory traffic and computation costs. The key characteristics of \OURL are: 
\begin{enumerate}
    \item It does not require fine-tuning.
    \item It quantizes \textit{all} weights and activations into 16-entry dictionaries of \textit{16 bit fixed-point} centroids storing \textit{all} weights and activations as narrow indexes in memory.
    \item It performs most computations directly on  4b dictionary indexes \textit{without} having to map them to their corresponding 16b centroids. This is \OURL's most innovative aspect.
    \item It uses fixed-point arithmetic yielding 16b fixed-point activations. 
    \item It quantizes the 16b fixed-point output activations into 4b indexes on-the-fly.
    \item It generates all dictionaries \textit{off-line} using a light-weight method utilizing linear transformations over a pre-generated, model independent ``Golden Dictionary''. 
    \item The Golden Dictionary is symmetric around zero requiring only half of the entries to be stored.
    \item It uses two dictionaries per tensor, one for the bulk of the values and another for outliers.
    \item Optionally, \OURL can be used as a compression plug-in to \textit{any} other accelerator.
\end{enumerate}
In more detail, \OURL exploits the skewed, bell-shaped (Gaussian) distribution of values naturally occurring in transformer models: most of values are densely populated around their mean (in a small range) and a small fraction of values (covering a wider range) are outliers.

A key innovation in \OURL is that it adjusts the centroids in each dictionary to fit the original tensor's Gaussian distribution \textit{and} an \textit{exponential function}. This is only possible because, for the limited range of values used in transformers, a Gaussian distribution can be approximated well by an exponential function. It is this property that allows \OURL to replace 16- or 32-bit floating-point multiply-accumulate operations with narrow, 3-bit, integer additions (see Section~\ref{sec:method}).

Additionally, \textit{all} the dictionaries are derived from a reference ``Golden'' Dictionary which is created using a non-linear quantization scheme based on Agglomerative clustering~\cite{agglo}. The Golden Dictionary is independent of specific models and layers and the quantization method is shared across all models. This results in a lightweight post-training quantization process that simply applies a linear transformation of the Golden Dictionary so that it fits the distribution of each tensor (as described by the mean and standard deviation).

Optionally, \OURL can be used just for memory compression to only compress values to 4b indexes when storing them to higher and more energy demanding levels of the memory hierarchy while expanding them to 16b fixed-point or FP16 floating-values when requested by the lower memory hierarchy levels or directly by the processing units. 

We evaluate \OURL over several FP16 pre-trained NLP models and highlight the following experimental findings:

\begin{itemize}
    \item \OURL quantizes model parameters (weights, embeddings) and activations to 4b indexes with an average of 1.5\% outliers for parameters and 4.5\% outliers for activations while maintaining model accuracy.
    \item \OURL's compute units consume $2.7\times$ less energy and are $1.6\times$ smaller than FP16 Tensor Cores units and boost energy efficiency by one to two orders of magnitude based on different configurations.  
    \item \OURL as a memory-compression-only method for FP16 models when deployed over a Tensor-cores based accelerator magnifies on-chip capacity and reduces off-chip traffic by $4\times$ improving overall energy efficiency by $11\times$.
\end{itemize}

\section{\OURLCORE\ Quantization}\label{sec:method}

It has been repeatedly shown that the values flowing through state-of-the-art attention-based NLP models form bell-shaped distributions per layer\cite{G_DNN,gordon2020compressing,GOBO}.

\OURL's quantization exploits these naturally occurring distributions that \textit{all} weights and \textit{most} activations exhibit. In contrast to past quantization methods that quantize each tensor separately, \OURL uses a ``Golden Dictionary'' approach which proceeds as follows:

\noindent\textbf{Step 1: Golden Dictionary Generation: }\OURL first generates and quantizes randomly generated, bell-shaped distributions with a mean of zero and a standard deviation of one. This step is performed once and is independent of the model under consideration. The resulting dictionary of centroids becomes \OURL's ``Golden Dictionary''. In this work, this dictionary contains 16b fixed-point values. 

\noindent\textbf{Step 2: Per Tensor (Activation and Weights) Dictionary Generation: }\OURL performs \textit{a profiling run} of the model collecting samples of the activation tensors. It uses the activation tensor samples and the statically known weight tensors to scale and shift the Golden Dictionary to best fit each tensor. These linear transformations utilize the target tensor's mean and standard deviation. \OURL produces two dictionary sets per layer, one per input tensor --- depending on the layer the input is a (weight, activation) or an (activation, activation) tensor pair. Per tensor, \OURL splits the value range into two subsets: 1)~the Gaussian subset (G) contains the majority of values (more than 98\% for weights and 95\% for activations) which covers values near the mean. 2) The Outliers (OT) which are the rest of the values that are scattered in a much wider range. 
\OURL's profiling-based approach for activation works well; while individual activation values vary with the input, their overall distribution per layer remains relatively unaffected. To validate this approach profiling runs use a single randomly selected batch containing 8 input samples (however, runs with even fewer input samples proved enough). Efficacy measurement runs use a non-overlapping set of input samples from the validation set of each target task. While prior quantization methods use an iterative per tensor approach~\cite{han2015deep,GOBO,q-bert}, \OURL's quantization method, does so only during the Golden Dictionary generation. Per tensor dictionary generation is non-iterative, resulting in a much faster overall process.

\noindent\textbf{Step 3: Pre-Encoding of Weights: } \OURL then encodes all weight tensors as indexes to their dictionaries (done offline). 

At this point, the model is ready for inference.

\subsection{Inference: Runtime Encoding/Decoding of Activations } During runtime \OURL produces 16b fixed-point output activations. These are converted to short indexes to the respective dictionaries prior to storing to memory. Since the dictionaries have few entries, this is a lightweight process that requires only a few runtime comparisons. Recall that in the original model, each output activation typically entails several hundreds of multiply-accumulate operations. As Section~\ref{sec:exp:approx} will explain, the \OURL accelerator performs most of the computation directly on indexes and does not need to convert input activations into their 16b centroids.

The rest of this section explains the dictionary generation process and the exponential function approximation used to replace the original floating-point multiply-accumulate operations with narrow fixed-point operations comprising mostly additions of dictionary indexes.

\subsection{Generating the Golden Dictionary (GD)}\label{sec:gd:gen}
\begin{figure}[t!]
\centering
\includegraphics[width=0.45\textwidth]{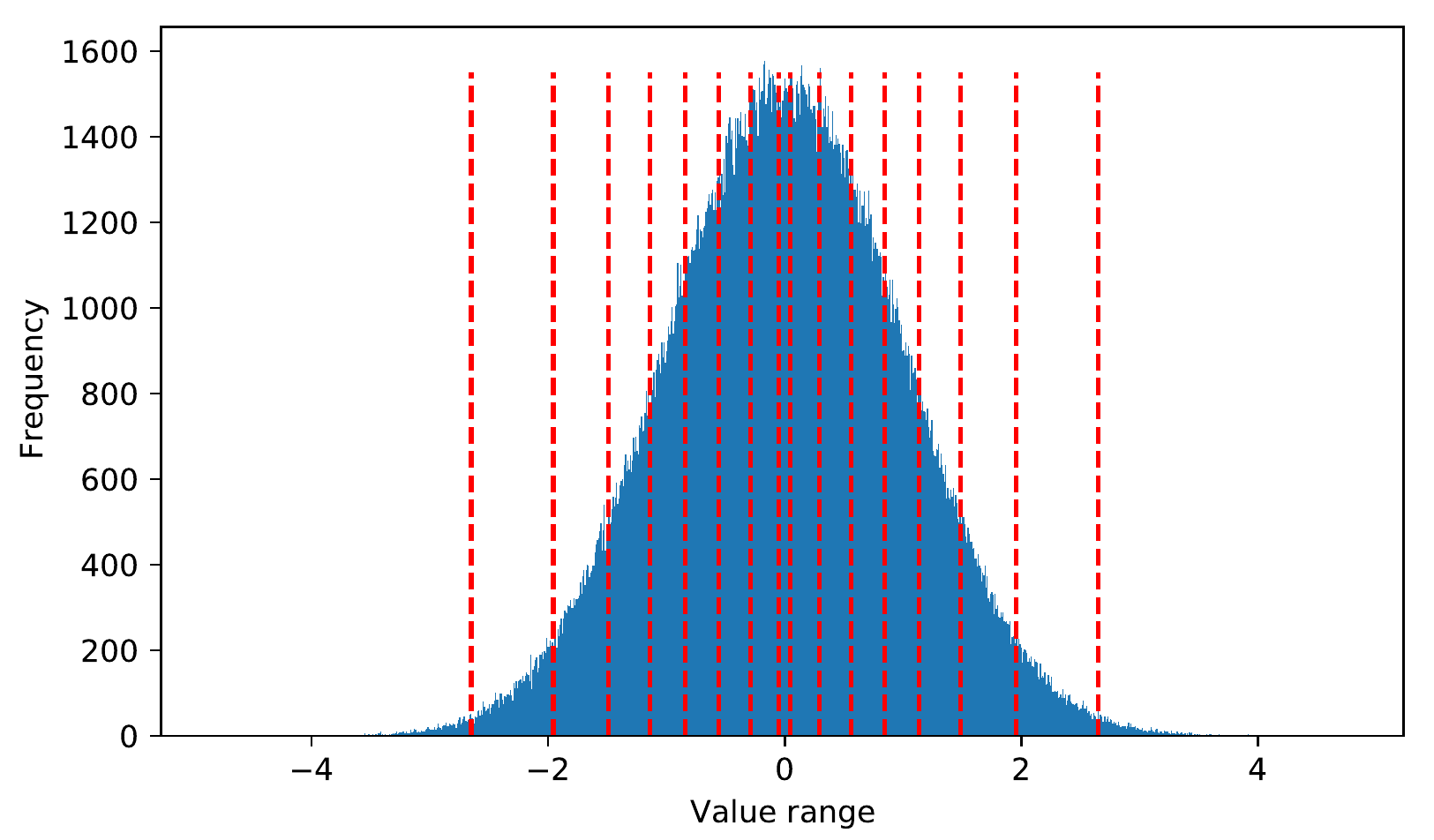}
\caption{Generating Golden Dictionary from a Random Gaussian Distribution Using Agglomerative Clustering.}
\label{fig:AggloQuant}
\end{figure}

The core of \OURL's dictionary generation method is Agglomerative Clustering (AC), a bottom-up approach which initially considers each value  as a separate cluster and that proceeds to iteratively merge the closest clusters to reduce overall cluster count to a desired target value. In contrast to K-means, the method of choice in several prior works, Agglomerative Clustering is not affected by the initial cluster selection and results in higher accuracy in the quantized model. However, applying AC even when done offline given that weight or activation tensors can contain nearly a million values is impractical since AC requires $\mathcal{O}(n^2)$ memory and $\mathcal{O}(n^3)$ runtime. 
\OURL overcomes this challenge by exploiting the morphological similarity of the value distributions across tensors to apply AC using a ``Golden Dictionary'' which it then scales to best fit each individual tensor.
In essence, this approach applies AC run on a representative distribution to \textit{estimate} AC's clustering for the per tensor distributions. The process is simple: first, generate a random Gaussian distribution with 50,000 samples with a mean of zero ($m=0$) and a standard deviation of one ($s=1$). Then apply AC method on this distribution to produce the quantization dictionary. To create the Golden Dictionary, we repeat this process and compute an average over quantization dictionaries. Figure~\ref{fig:AggloQuant} shows the histogram for an example generated distribution and the resulting Golden Dictionary. In this work, Golden Dictionary is generated once and is reused across all models~studied. 

\subsection{Generating Per Tensor Dictionaries}\label{sec:tensord:gen}
\OURL fits the Golden Dictionary (GD) to each tensor by first determining the mean ($m$) and the standard deviation ($s$) of the tensor's values. These two parameters are constant for weights and are estimated using profiling for activations. A simple linear transformation of GD is all that is needed:  $GD \times s +m $. This is done once per tensor using profiling.

Dictionary size affects overall accuracy. In general, the more entries a dictionary has, the better it represents the original tensor distribution and the better it preserves overall model accuracy. However, larger dictionaries incur higher overheads.
To better balance dictionary overheads while preserving accuracy \OURL uses two small dictionaries per tensor. The first ``Gaussian'' dictionary quantizes the bulk of the values, whereas the second ``Outlier'' dictionary maps the few remaining values of much larger magnitude. To best explain how \OURL chooses the respective ranges, it is best to first discuss how \OURL manages to map what were originally floating-point values and operations into fixed-point values and operations. Unique to \OURL, the bulk of these operations manipulate narrow fixed-point dictionary \textit{indexes} instead of their respective \textit{centroids}.

\begin{figure}[t!]
\centering
\includegraphics[width=0.35\textwidth]{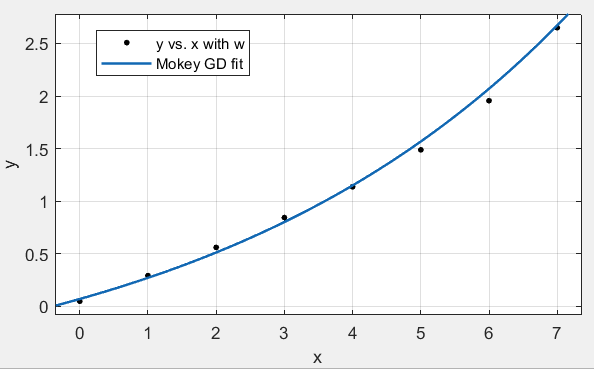}
\caption{Fitting an Exponential Curve to Golden Dictionary.}
\label{fig:MatlabF}
\end{figure}

\subsection{Integer Exponents for Compute Acceleration}\label{sec:exp:approx}

Mokey quantizes weights and activations to a few unique values. For clarity, this section assumes that there is a single dictionary of 16 values which all follow a Gaussian distribution with a narrow range near its mean (outlier handling is described in Section~\ref{sec:outliers}). With this quantization scheme, there are just 256 ($16 \times 16$) possible multiplication products instead of the $2^{32}$ possible products with the original FP16 model. This represents an opportunity to replace the native FP16 multiply-accumulate (MAC) operations, with counting per possible product (integer histogram), followed by a final multiplication of the integer counts with the product of their respective centroids. However, we anticipate that having 256 counters per processing element will be impractical due to high energy and latency overhead compared to an FP16 MAC~unit.

\begin{figure}[t!]
\centering
\includegraphics[width=0.49\textwidth]{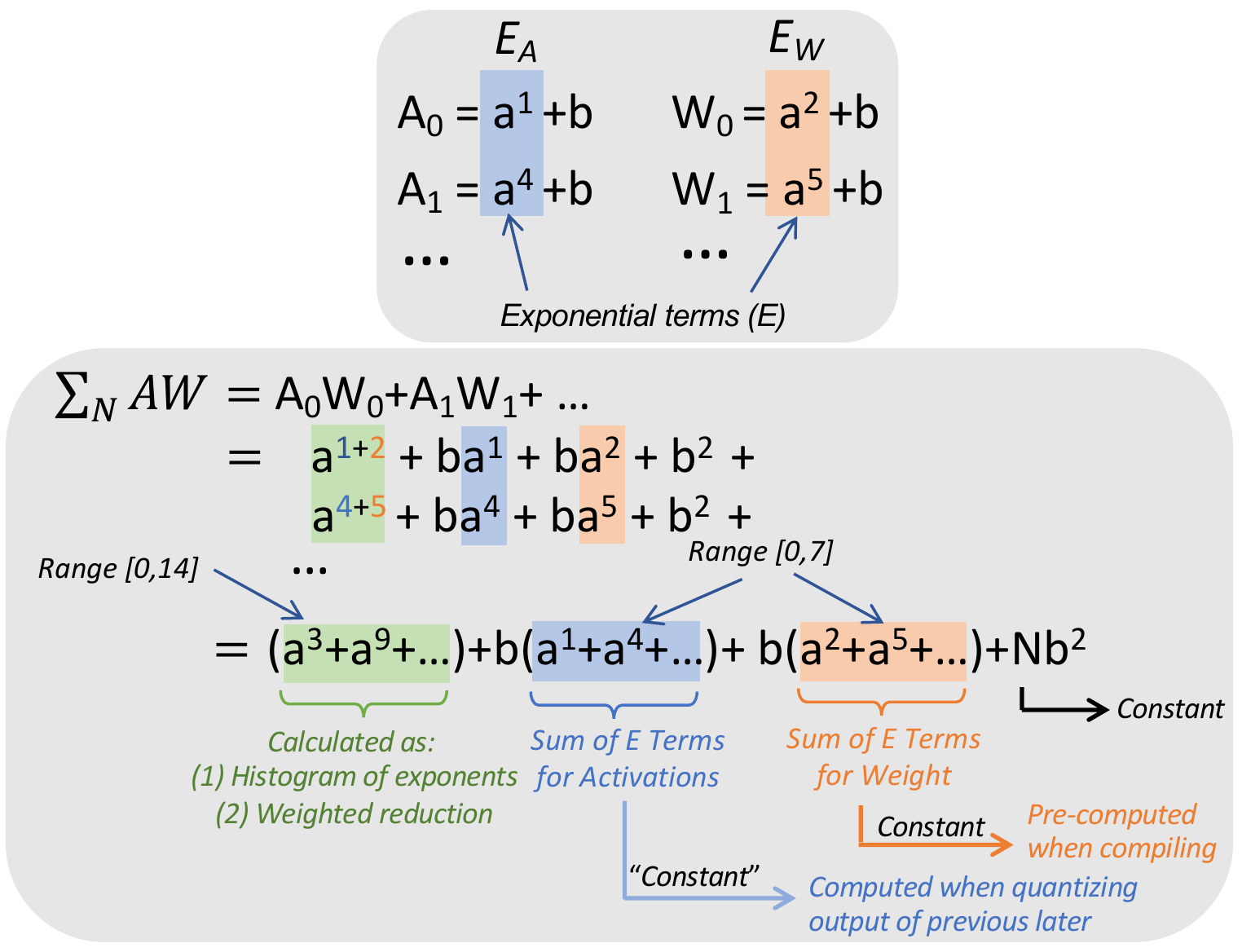}
\caption{{Calculating output activations without having to expand the quantized input activations and weights to their corresponding centroids.}}
\label{fig:Sec2Fig}
\end{figure}

We observe that a Gaussian distribution, provided that we are only interested in values near its mean, can be approximated well with an exponential function where values are represented as $a^{int}+b$, where $int$ is an integer index and $a$ and $b$ are appropriately chosen constants. As we explain here, this approximation enables \OURL to reduce what otherwise would have been 256 possible products into just 15 possible exponent sums, making a histogram-based processing MAC unit practical. {Figure}~\ref{fig:Sec2Fig} illustrates, informally, how \OURL decomposes MAC operations into exponent counting (histogram) for quantized indexes and some constants. In this simplified example, all values are assumed to be positive and the layer's mean and standard deviation are assumed to be zero and one, respectively. 

In more detail: Since the Golden Dictionary is symmetrical around $0$ and for a limited range, we first fit an exponential curve to one-half of the Golden Dictionary. The other half holds the opposite values. We used MATLAB's curve-fitting toolbox~\cite{MATLAB:2020} to fit a curve of the form $a^{int}+b$ where $a, b$ are the parameters that we are fitting and $int$ is a list of integers from zero to half of the dictionary size. Since we have more values near zero and the density of the values decreases as we go sideways, we apply a weighting scheme during curve fitting to emphasize more densely populated ranges. The weighting scheme uses a unit weight for the outer bin, and doubles the weight for the bins as we move towards zero.
For 4 bit quantization, this results in 16 bins which are symmetrical around zero: 8 are positive and 8 are negative with identical magnitudes as their corresponding positive bin. We fit the $GD = a^{int}+b$ curve on these 8 positive values where $a=1.179$, $b=-0.977$, where $int$ is an integer in range of $[0,7]$ and the fitting weights are in $[2^7 , 2^0]$ range. 
Figure~\ref{fig:MatlabF} shows the resulting exponential function (curve) and the original Golden Dictionary entries (dots).

In the resulting 4 bit quantization, the most significant bit is the sign of the value (0: positive , 1: negative) and the following 3 bits are the index. For instance, if the quantized representation is $0b1011$ the corresponding value would be $\mathbf{-}(a^{\mathbf{3}}+b)$. We emphasize that $a$ and $b$ are fitted to the Golden Dictionary and this process is performed only \textit{once} prior to scaling the resulting dictionary to each tensor. 

\noindent\textbf{Computing using Dictionary Indexes instead of Dictionary Centroid Values:} We can now exploit that our values fall on an exponential curve to simplify multiplication. At a high-level the method takes advantage of property $a^i\times a^j=a^{i+j}$. After approximating the dictionary with values on an exponential curve, we are guaranteed that when we are multiplying two input tensors, input values are of the form $\theta(a^{int}+b)\times s + m$ where $theta$ is the sign, $int$ a 3b Golden Dictionary index, and $s$ and $m$ are constants, respectively, the standard deviation and the mean of the tensor.

For clarity, let us assume that the layer operates on a weight $W$ and an activation $A$ tensor. The respective values would then be of the form:
$\theta_W(a^{int_W}+b)\times s_W + m_W$ and  $\theta_A(a^{int_A}+b)\times s_A + m_A$ where $\theta_A, \theta_W$ are signs, $int_W, int_A$ are integer indexes, $s_W, s_A$ are standard deviations and $ m_W ,m_A$ are means for the corresponding weight and activation tensors, respectively. 

The weight and activation values can be simplified to:
\vspace{-4pt}
\begin{dmath}[style={\scriptsize},compact, spread={-1pt} ]
A={{\theta }_{A}}({{a}^{in{{t}_{A}}}}+b)\times {{s}_{A}}+{{m}_{A}}={{\theta }_{A}}{{s}_{A}}.{{a}^{in{{t}_{A}}}}\underbrace{+{{\theta }_{A}}{{s}_{A}}.b+{{m}_{A}}}_{{{\mu }_{A}}}={{\theta }_{A}}{{s}_{A}}.{{a}^{in{{t}_{A}}}}+{{\mu }_{A}} \label{eq:A} \\ 
\end{dmath}

\vspace{-10pt}\begin{dmath}[style={\scriptsize},compact, spread={-1pt}]
W={{\theta }_{W}}({{a}^{in{{t}_{W}}}}+b)\times {{s}_{W}}+{{m}_{W}}={{\theta }_{W}}{{s}_{W}}.{{a}^{in{{t}_{W}}}}\underbrace{+{{\theta }_{W}}{{s}_{W}}.b+{{m}_{W}}}_{{{\mu }_{W}}}={{\theta }_{W}}{{s}_{W}}.{{a}^{in{{t}_{W}}}}+{{\mu }_{W}}  \label{eq:W}
\end{dmath}

where $a$ is a fitted parameter which is constant for all tensors across all layers and models, a sign ($\theta$), $int_A$ and $int_W$ are dictionary indexes, $S_A$ and $S_W$ are per tensor scalars scaling factors,  and $\mu_A$ and $\mu_W$ are shifts.  

Each output activation is a sum of activation times weight products, which given the form of the input values can be expanded into a sum of four terms:

\vspace{-5pt}\begin{dmath}[style={\scriptsize},compact, spread={-1pt}]
   \sum\limits_{i=0}^{n-1}{{{A}_{i}}.{{W}_{i}}}=\sum\limits_{i=0}^{n-1}{\left( {{\theta }_{{{A}_{i}}}}{{s}_{A}}.{{a}^{{{\operatorname{int}}_{{{A}_{i}}}}}}+{{\mu }_{{{A}_{i}}}} \right)\left( {{\theta }_{{{W}_{i}}}}{{s}_{W}}.{{a}^{{{\operatorname{int}}_{{{W}_{i}}}}}}+{{\mu }_{{{W}_{i}}}} \right)} \nonumber \\ 
  =\underbrace{{{s}_{A}}.{{s}_{W}}\sum\limits_{i=0}^{n-1}{\left( {{\theta }_{{{A}_{i}}}}{{\theta }_{{{W}_{i}}}}{{a}^{{{\operatorname{int}}_{{{A}_{i}}}}+{{\operatorname{int}}_{{{W}_{i}}}}}} \right)}}_{SoI}+\underbrace{{{s}_{A}}\sum\limits_{i=0}^{n-1}{\left( {{\mu }_{{{W}_{i}}}}{{\theta }_{{{A}_{i}}}}{{a}^{{{\operatorname{int}}_{{{A}_{i}}}}}} \right)}}_{SoA}+\underbrace{{{s}_{W}}\sum\limits_{i=0}^{n-1}{\left( {{\mu }_{{{A}_{i}}}}{{\theta }_{{{W}_{i}}}}{{a}^{{{\operatorname{int}}_{{{W}_{i}}}}}} \right)}}_{SoW}+\underbrace{\sum\limits_{i=0}^{n-1}{\left( {{\mu }_{{{A}_{i}}}}.{{\mu }_{{{W}_{i}}}} \right)}}_{PoM} \label{EQ:OutputAct}
\end{dmath}

The first term ($SoI$) is the sum of ${{a}^{{{\operatorname{int}}_{{{A}_{i}}}}+{{\operatorname{int}}_{{{W}_{i}}}}}}$. Assuming 4b quantization (3b indexes plus 1b signs), $int_A+int_W$ will be in range of $[0+0 , 7+7]= [0,14]$ which given the quantization contains 15 unique values. Since the exponents are limited to 15 values we can first add $int_A$ and $int_W$ and count how many times each exponent occurs (count how many $a^0$, $a^1$...$a^{14}$ we have overall) then with 15 MAC operations (occurrences $\times a^0 +...+$occurrences $\times a^{14}$) we can compute the $SoI$ term.

The second term ($SoA$) can be decomposed into two terms as per Eq.~\ref{EQ:SOA}. The first term ($\mathit{SoA}_1$) is a summation of activations with respect to the sign of both weight and activation. This term is computed during computation similar to $\mathit{SoA}$, but it needs smaller counters as the range of activation is limited to $[0,7]$. The second term ($\mathit{SoA}_2$) is the sum of the input activations for this layer. This term can be computed while the output of previous layer is being quantized so that is available in advance of this layer.:

\vspace{-5pt}\begin{dmath}[style={\scriptsize},compact, spread={-1pt}]
 SoA={{s}_{A}}\sum\limits_{i=0}^{n-1}{\left( {{\mu }_{{{W}_{i}}}}{{\theta }_{{{A}_{i}}}}{{a}^{{{\operatorname{int}}_{{{A}_{i}}}}}} \right)}={{s}_{A}}\sum\limits_{i=0}^{n-1}{\left( ({{\theta }_{{{W}_{i}}}}{{s}_{W}}.b+{{m}_{W}}){{\theta }_{{{A}_{i}}}}{{a}^{{{\operatorname{int}}_{{{A}_{i}}}}}} \right)}  \nonumber  \\ 
 =\underbrace{{{s}_{A}}.{{s}_{W}}.b\sum\limits_{i=0}^{n-1}{\left( {{\theta }_{{{W}_{i}}}}{{\theta }_{{{A}_{i}}}}{{a}^{{{\operatorname{int}}_{{{A}_{i}}}}}} \right)}}_{So{{A}_{1}}}+\underbrace{{{s}_{A}}.{{m}_{W.}}\sum\limits_{i=0}^{n-1}{\left( {{\theta }_{{{A}_{i}}}}{{a}^{{{\operatorname{int}}_{{{A}_{i}}}}}} \right)}}_{So{{A}_{2}}}  \label{EQ:SOA}
\end{dmath}
\vspace{-5pt}
The third term ($SoW$), similar to $SoA$ is decomposed into two term as per Eq~\ref{EQ:SoW}. The first term ($SoW_1$) is the summation of the weights with respect to the sign of both weight and activation, and it is computed similar to ($SoA_1$). The second term ($SoW_2$) is a summation over weights and is a constant that can be computed statically before inference.

\vspace{-5pt}\begin{dmath}[style={\scriptsize},compact , spread={-1pt}]
 SoW={{s}_{W}}\sum\limits_{i=0}^{n-1}{\left( {{\mu }_{{{A}_{i}}}}{{\theta }_{{{W}_{i}}}}{{a}^{{{\operatorname{int}}_{{{W}_{i}}}}}} \right)}={{s}_{W}}\sum\limits_{i=0}^{n-1}{\left( ({{\theta }_{{{A}_{i}}}}{{s}_{A}}.b+{{m}_{A}}){{\theta }_{{{W}_{i}}}}{{a}^{{{\operatorname{int}}_{{{W}_{i}}}}}} \right)}  \nonumber  \\ 
 =\underbrace{{{s}_{W}}.{{s}_{A}}.b\sum\limits_{i=0}^{n-1}{\left( {{\theta }_{{{A}_{i}}}}{{\theta }_{{{W}_{i}}}}{{a}^{{{\operatorname{int}}_{{{W}_{i}}}}}} \right)}}_{So{{W}_{1}}}+\underbrace{{{s}_{W}}.{{m}_{A}}.\sum\limits_{i=0}^{n-1}{\left( {{\theta }_{{{W}_{i}}}}{{a}^{{{\operatorname{int}}_{{{W}_{i}}}}}} \right)}}_{So{{W}_{2}}} \label{EQ:SoW}
\end{dmath}
\vspace{-5pt}
The last term ($PoM$) is decomposed in Eq.~\ref{EQ:POM} to 4 terms. The first term is a summation of the sign of multiplication ($PoM_1$) which is computed during inference. The second and the third terms are the summation of activations' sign (known when quantizing last layer) and weights' sign (known before inference) and the last term is a constant.

\vspace{-5pt}\begin{dmath}[style={\scriptsize},compact , spread={-1pt}]
  PoM=\sum\limits_{i=0}^{n-1}{\left( {{\mu }_{{{A}_{i}}}}.{{\mu }_{{{W}_{i}}}} \right)}=\sum\limits_{i=0}^{n-1}{\left( \left( {{\theta }_{{{A}_{i}}}}{{s}_{A}}.b+{{m}_{A}} \right)\left( {{\theta }_{{{W}_{i}}}}{{s}_{W}}.b+{{m}_{W}} \right) \right)} \nonumber  \\ 
 =\underbrace{{{s}_{A}}.{{s}_{W}}.{{b}^{2}}\sum\limits_{i=0}^{n-1}{\left( {{\theta }_{{{A}_{i}}}}{{\theta }_{{{W}_{i}}}} \right)}}_{Po{{M}_{1}}}+\underbrace{{{s}_{A}}.{{m}_{W}}.b\sum\limits_{i=0}^{n-1}{\left( {{\theta }_{{{A}_{i}}}} \right)}}_{Po{{M}_{2}}}+\underbrace{{{s}_{W}}.{{m}_{A}}.b\sum\limits_{i=0}^{n-1}{\left( {{\theta }_{{{W}_{i}}}} \right)}}_{Po{{M}_{3}}}+\underbrace{n.{{m}_{A}}.{{m}_{W}}}_{Po{{M}_{4}}}  \label{EQ:POM}
\end{dmath}
\vspace{-5pt}
In conclusion computation can proceed as follows: 
\begin{enumerate*}
    \item Add the indexes (3 bit values) for a weight and an activation we want to multiply.
    \item XOR the sign bit of weight and activation and based on that result, 
    \item increment or decrement the occurrences tables for $SoI$ (15 entries), $SoA_1$ (8 entries), $SoW_1$ (8 entries), and $PoM_1$ (1 entry).
    \item After filling the occurrences tables (which typically will require thousands of cycles given the tensor sizes), multiply each count with its corresponding value ($a^0 ... a^{15}$) and accumulate all products into a single value.
    \item The final values are multiplied by their coefficients and added with pre-computed terms (which at this point are constants) ($SoA_2,SoW_2,PoM_{2,3,4}$) producing the final output activation. 
\end{enumerate*}

The output activation can then be quantized by comparing it against the dictionary for the output activation tensor. In practice, we find that 16-entry dictionaries prove sufficient. Values are quantized to a 4b indexes which are a 3b index into the stored dictionary entries and a 1b sign.

\subsection{Outliers}
\label{sec:outliers}
Outliers (OT) are exceedingly rare and represent less than 2\% of weights and 5\% of activations (varies per model and task). However, they do span over a disproportionally large value range. To quantize outliers, we adapt the Golden Dictionary approach. 

We could na\"ively use a common dictionary for all values. Starting with the 16 centroid dictionary for the G values, we found that to support outliers, we need to widen the index range to $int=45$. Encoding the index to such a dictionary would require 6(index)+1(sign)=7 bits per value.
We use a two-prong approach to limit index lengths to 4b. First, we use a separate outlier dictionary. Second, we take advantage of the exceedingly rare occurrence of outliers to store all values, outlier or Gaussian, using a 4b index. A separate, space- and access-efficient list of pointers explicitly identifies which indexes correspond to outliers (all others index to the Gaussian dictionary). Section~\ref{sec:mem:layout} describes the encoding.

Outliers require different handling during computation as they do not fit under the same exponents (0 to 7) as the Gaussian values. For any product where at least one of the values is an outlier, \OURL converts the values into their centroids and performs a multiply-accumulate. Since outliers are exceedingly rare, this is a rare occurrence (less than 4\% of the multiplications in BERT). 
However, unless further action is taken, the centroids would be FP16 numbers necessitating FP16 capable MAC hardware. Fortunately, \OURL is further enhanced to require only fixed-point values and hardware. 

\subsection{Integer Computation Throughout}
Integer compute units are faster and more energy-efficient than floating-point units. \OURL maps what were originally floating-point computations to the fixed-point (integer) domain. 
The conversion to fixed-point arithmetic is done during profiling. The process starts after the mean, standard deviation, the dictionary (bins) that we are going to use for outliers and the few constants (SoW, PoM) have been derived. To convert these parameters to integer, \OURL first needs to set the total bit-width ($b$) and to compute the number of fractional bits per layer ($frac$). The number of fractional bits ($frac$) needed is:  
\begin{equation}
\label{eq:frac1}
    frac=b-\text{ceil}\left( {{\log }_{2}}(\max -\min ) \right )
\end{equation}
where $max$ and $min$ are the maximum and minimum values that appear for the layer. The float ($fl$) to fixed-point($fx$) mapping then is:
\begin{equation}
\label{EQ:fixpt}
    {fx} = \operatorname {round}(fl \times 2^{frac})/2^{frac}
\end{equation}

\subsection{Summary}
In summary,  \OURL quantized an input model as follows: First, it generates the Golden Dictionary for quantization. This step is done once and the output is reused across \textit{all} models. Second, it quantizes the model's parameters and embeddings. On average, these statically known values consist of about 98.5\% G values and 1.5\% outliers. G values are stored in the form of exponents (4-bit) and outliers are stored as small (16 entry--4-bit) dictionary indexes where the dictionary entries are 16b fixed-point values. A list of pointers identifies which indexes are to the outlier dictionary. Third, \OURL performs a run over a randomly selected batch of inputs to generate the dictionaries for the activations. The profiling is to find the mean, standard deviation, and the outlier bins that activations are using. The dictionaries and other constants per tensor are stored along with the model. The space needed for this metadata pales in comparison with the size of the respective tensors; a G dictionary (8$\times$16b entries), an OT dictionary (16$\times$16b), plus a few constants and a list of outlier pointers.

\section{\OURLCORE\ Memory Layout and Processing Units}

This section describes the memory encoding and processing units of \OURL. Combined they reduce memory traffic and footprint while performing most computations using narrow fixed-point arithmetic. 

\begin{figure}[t!]
\centering
\includegraphics[width=0.35\textwidth]{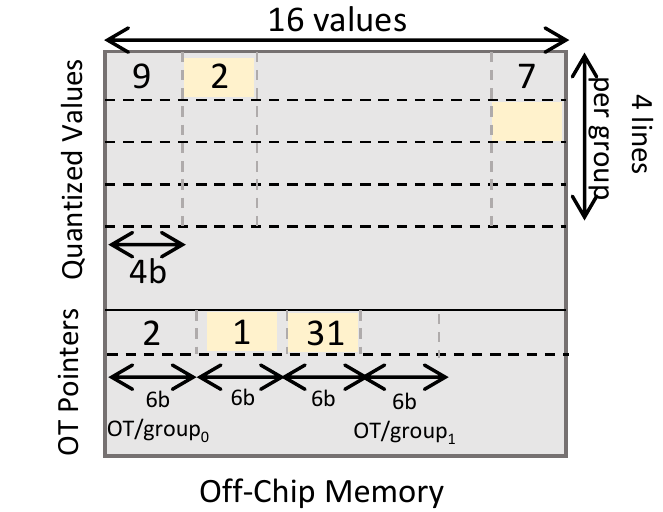}
\caption{DRAM layout}
\label{fig:DRAM_layout}
\vspace{-10pt}
\end{figure}

\subsection{Off-Chip Memory Layout}\label{sec:mem:layout}
To take advantage of \OURL's quantization for reducing off-chip memory traffic and footprint, we need to map Gaussian and outlier values in a DRAM-friendly data container. Per tensor, Gaussian and outlier values are represented as 4b indexes. Since outliers are rare, \OURL stores all values using 4b indexes as shown in Figure~\ref{fig:DRAM_layout}. An extra list of pointers identifies which of those indexes are outliers. The pointers are encoded as follows:  Conceptually, the regular ``Quantized value'' 4b index array is split into groups of 64, 4b indexes. To identify those indexes that are outliers, the ``OT Pointers'' list first stores outlier count per group, followed by a list of 6b indexes marking their relative position within the group. In the example shown, $group_0$ has two outliers at positions 1 and 31. This container is DRAM-friendly as accesses to the two sub-arrays will be done through two streams, each proceeding through the respective areas sequentially with the bulk of accesses going to the ``Quantized Values'' area. For simplicity, at an appropriate level of the on-chip hierarchy the values can be expanded to 5b (dictionary selection/1b, sign/1b, centroid index/3b) indexes. This facilitates single stream accesses per tensor.

\subsection{Processing Units}\label{sec:pe:hw}
\OURL hardware accelerator enables compute units to directly operate on the quantized indexes using narrow fixed-point processing elements (PEs) for the bulk of computation (Gaussian ``G'' values). Only for products where at least one of the operands (weight or activation) is an outlier (``OT''), \OURL PEs convert the values into their respective int16 centroids. This requires a simple lookup to the respective dictionary.

\begin{figure*}
\centering
\includegraphics[width=0.95\textwidth]{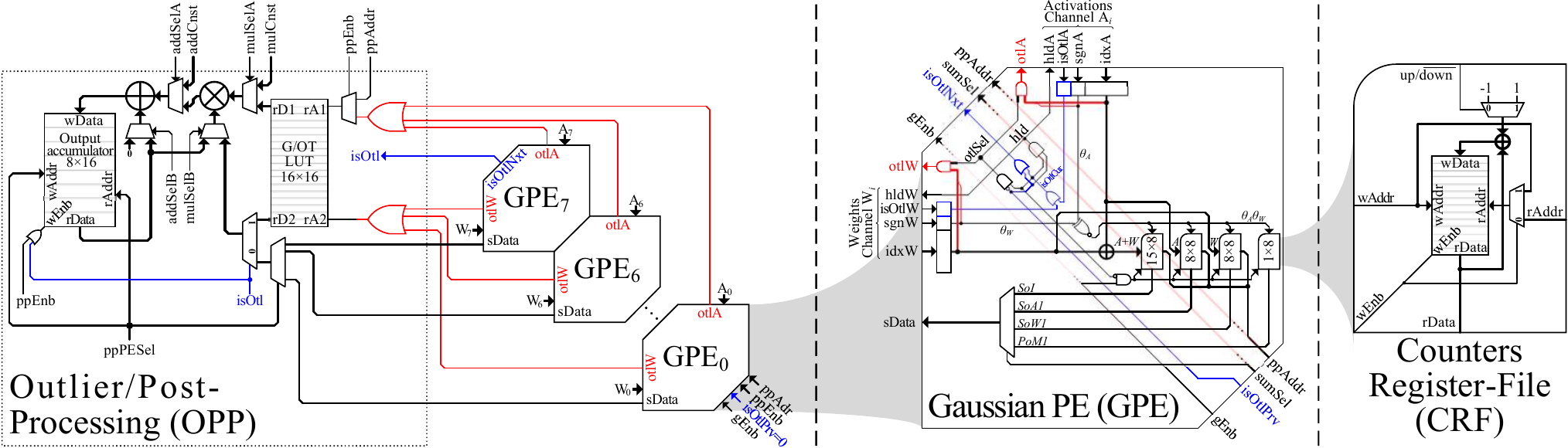}
\caption{Architecture of \OURL Hardware Accelerator.}
\label{fig:Mokey_tile}
\end{figure*}

\begin{figure}[t!]
\centering
\includegraphics[width=0.43\textwidth]{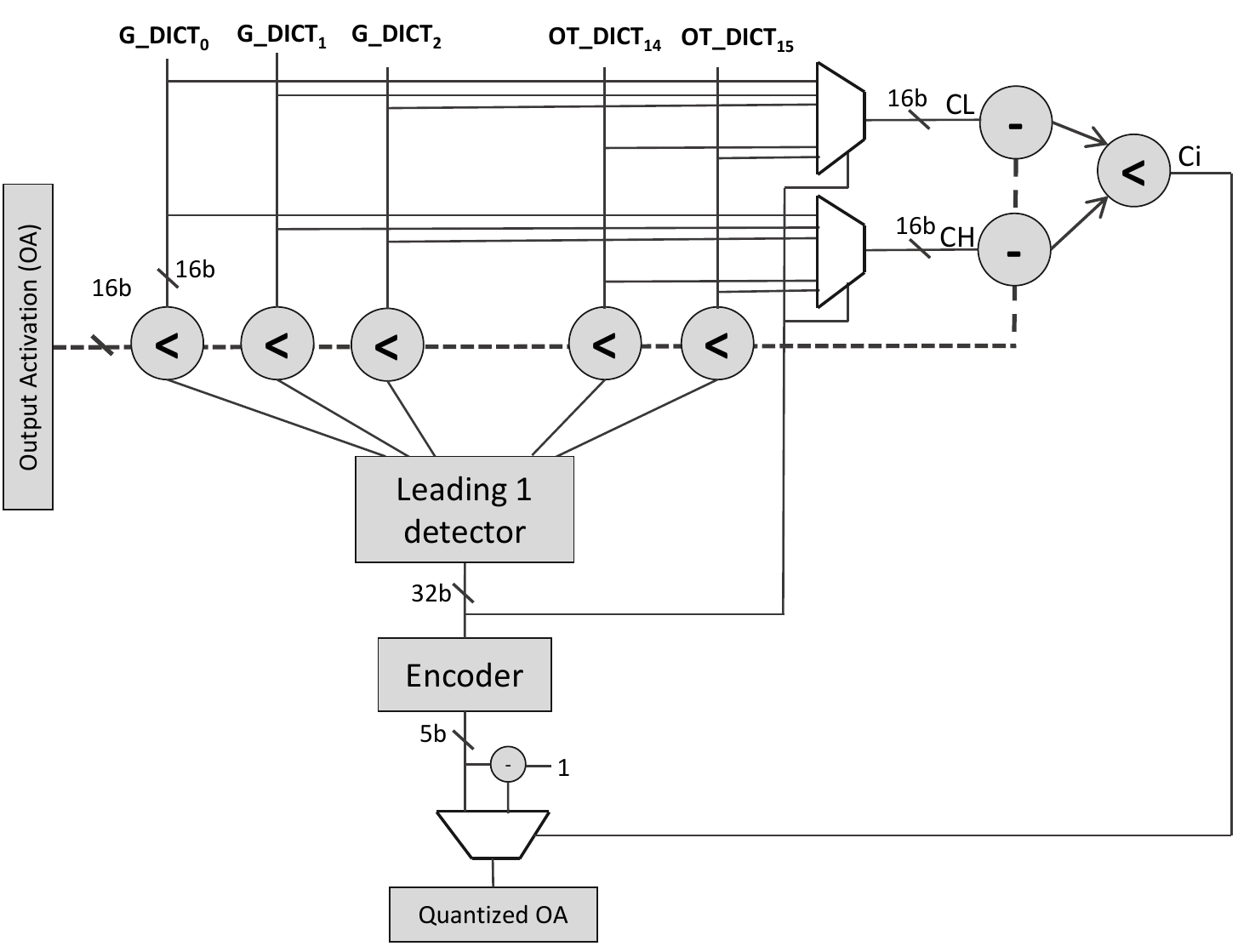}
\caption{Output Quantization Engine.}
\label{fig:out_quant}
\end{figure}

\noindent\textbf{Tile architecture: }
\autoref{fig:Mokey_tile} shows the architecture of processing unit of the \OURL hardware accelerator. As shown on the left, the accelerator comprises an array of 8 cascaded Gaussian PEs (GPEs) which share an outlier/post-processing circuitry (OPP). At peak, each GPE$_i$ processes a group of compressed activations ($A_{0\cdots7}$) and the corresponding group of compressed weights ($W_{0\cdots7}$). The GPEs process the Gaussian-distributed inputs only; the outliers are sent to the OPP unit for lookup and accumulation. The OPP unit is activated only when an outlier is received, or by the end of the computation for post-processing. At a high-level, most activation and weight pairs will use Gaussian values and will be processed by the GPEs that accumulate the partial sums described in Section~\ref{sec:exp:approx}. The occasional pairs containing outliers are processed in the OPP. Since these are rare, they are processed one at a time for simplicity and area-efficiency. After the full tensors have been processed, a final step uses the OPP to accumulate and scale the various summations into the output 16b activation. The unit processes activations and values that have been encoded in 5b (1b dictionary index, 1b sign, 3b dictionary index).

The GPE is detailed in \autoref{fig:Mokey_tile} (middle). In the core of each GPE, four individually sized Counter Register-Files (CRFs) accumulate the $SoI$, $SoA_1$, $SoW_1$, and $PoM_1$ summations (Section~\ref{sec:exp:approx}). The organization of the CRF units is illustrated in \autoref{fig:Mokey_tile} (right). Each line (addressed by \texttt{wAddr}) can be incremented or decremented using \texttt{up/$\overline{\texttt{down}}$}, if \texttt{wEn} is asserted. The read address, \texttt{rAddr} is used to scan the register-file content during post-processing using the post-processing address \texttt{ppAddr}. The multiplication output sign ($\theta_A\theta_W\equiv\texttt{sgnA}\mathbin{\oplus}\texttt{sgnA}$) is used to increment/decrement all CRFs. $SoI$ is counted using a $15\times8$ CRF, and is addressed with $A+W\equiv\texttt{idxA+idxW}$. $SoA_1$ is counted using a $8\times8$ CRF, and is addressed with $A\equiv\texttt{idxA}$. $SoW_1$ is counted using a $8\times8$ CRF, and is addressed with $W\equiv\texttt{idxW}$. Finally, $PoM_1$ is counted using a $1\times8$ CRF (a single byte). As the post-processing is performed only once for each layer, its timing is less critical. Thus, the post processing is performed serially. Only one of the summations ($SoI$, $SoA_1$, $SoW_1$, and $PoM_1$) is selected for post processing as \texttt{sData}, and is selected by the control signal \texttt{sumSel}.

\textit{Scheduling outliers processing: } Since it is exceedingly rare but possible that more than one GPE receive an outlier simultaneously, the processing of outliers must be scheduled: the lowest index GPE that contains an outlier is selected for processing; all other GPEs with outliers send a hold signal back through the input channel (\texttt{hldA}, and \texttt{hldW}, to hold the activations and weights, respectively). Each GPE generates a local outlier-is-present signal $\texttt{isOtlCur} = \texttt{isOtlA} \vee \texttt{isOtlW}$. A serial, cascaded leading-one-detector across all GPEs detects the first GPE with outlier ($\texttt{isOtlCur}=1$). 
The selected ``leading'' GPE with an outlier asserts its \texttt{otlSel} signal ($\texttt{otlSel}=1$) forcing the \texttt{otlSel} signals to be deasserted ($\texttt{otlSel}=0$) for all other GPEs. The one-hot-encoded \texttt{otlSel} is used to route the corresponding indexes to the OPP (routing is done via per GPE AND gates and an OR level shared across all GPEs). 

\textit{The Outlier/Post-Processing (OPP) unit: } First, when an outlier has been detected, where the \texttt{isOtl} signal is asserted, the OPP uses its G/OT-LUT to retrieve the outlier value and performs a MAC operation. Second, in Post-processing where \texttt{sData} will hold one of the summations ($SoI$, $SoA_1$, $SoW_1$, and $PoM_1$). A single PE will be selected iteratively using the selector \texttt{pePESel}. The G/OT-LUT will operate as a G-LUT and produce the bases of the summations. Finally, the bases from the G-LUT and the summations from the GPEs will be multiplied and accumulated.

\noindent\textbf{Output Activation Quantizers: }
{Prior to storing them in memory, the output activations (OA) of each layer are quantized using the prepared OT and G dictionaries. Figure}~\ref{fig:out_quant} {shows the organization of the quantization units. An output activation OA is compared with every centroid from both the G and the OT dictionaries. Since the dictionary values are \textit{sorted}, and assuming 16 values per dictionary, the output of the comparators will be a 32b vector of 0s followed by 1s. A leading-one detector drives two 32-to-1 multiplexers, respectively, selecting the two corresponding 16b centroids CL and CH: the one corresponding to the leading 1 position and the one before it or the same if the position happens to be the first one. To determine the nearest to the OA centroid of the two, OA is subtracted from each of CL and CH, and the two differences are compared to find the smaller of the two identified by the signal Ci. The relative position of this centroid is then encoded as a 5b index.
If written off-chip, a controller packs them to the format of Figure}~\ref{fig:DRAM_layout}.

\subsection{Using \OURLCORE\ For Memory Compression Only}
Optionally, \OURL can be used just as a memory compression method where the values are transparently converted to fixed-point 16b or (FP16 if desired) when written or read from an appropriate level in the memory hierarchy. In this case, when reading values, lookup tables can convert the indexes into their corresponding centroids. Quantizers placed just before the memory level pack the activation values into indexes to dictionaries. When the compression targets an on-chip buffer, the activations are best encoded with 5b (1b dictionary OT or G, 1b sign, 3b index), whereas when writing to off-chip memory, the \OURL off-chip memory format would improve space efficiency. Using the less efficient 5b representation on-chip avoids the overheads of creating the OT Pointer metadata and the need for two access streams per tensor.

\section{Evaluation}\label{sec:eval}

\subsection{Model Task Performance}
We first evaluate the effect of \OURL quantization on \textit{task performance} over various models, tasks, and datasets. Task performance refers to how \textit{well} the model performs each task. The metric used is task-dependent. We show that \OURL maintains task performance without requiring fine-tuning. 

We applied \OURL on the following models: DeBERTa~\cite{he2021deberta}, a recent model from Microsoft that is in the top-ranked models on GLUE~\cite{glue} benchmark, on Facebook's RoBERTa~\cite{robetra} which is used widely used in sentiment analysis, and on models from Google's BERT family of models~\cite{BERT} which powers the search engine to show more related results in search queries. The pre-trained models were obtained from the Hugging Face Model hub~\cite{Wolf2019HuggingFacesTS}. The Golden Dictionary was created by generating a random normal distribution and applying Agglomerative Clustering using the SciKit-Learn library~\cite{scikit-learn}.
Our main benchmark to evaluate model accuracy is the MNLI task in the GLUE dataset, as it is the most sensitive task to quantization and has the most comprehensive set for language inference in the dataset~\cite{gordon2020compressing,GOBO}. Furthermore, MNLI trained models are publicly available. Where possible, however, we also study STS-B and SQuAD.  

\noindent\textbf{BERT: } Bidirectional Encoder Representations from Transformers (BERT)~\cite{BERT} is a well-known and widely used transformer model for NLP applications such as similarity detection, sentiment analysis, and question answering. We apply \OURL on BERT-Base (12 Encoders, 110M parameters) and BERT-Large (24 Encoders, 340M parameters) on MNLI (The Multi-Genre Natural Language Inference), STS-B (Semantic Textual Similarity Benchmark) from GLUE benchmark, and SQuADv1~\cite{squad} (Stanford Question Answering Dataset) task. 

\noindent\textbf{RoBERTa: } A Robustly Optimized BERT Pretraining Approach (RoBERTa)~\cite{robetra} shares the same architecture as BERT but benefits from an optimized training recipe that leads to higher accuracy. We evaluate \OURL on RoBERTa-Large using MNLI, STS-B and SQuAD v1 tasks. RoBERTa-Large has 24 Encoder layers and a total of 340M parameters.  

\noindent\textbf{DeBERTa: } Decoding-enhanced BERT with disentangled attention (DeBERTa)~\cite{he2021deberta} is a state-of-the-art transformer model that increases the accuracy of Transformers by decoupling the position of the tokens from their context embedding. We evaluate DeBERTa-XL (48 Encoder layers, 750M parameters) on the MNLI task since trained MNLI models are publicly available. 

\begin{table*}
\centering
\ra{0.9}
\caption{The Effect of \OURL Quantization on Task Performance (see text for definition.)}
\label{tbl:accuracy}
\begin{tabular}{@{}lllcccccccc@{}}\toprule[1pt]
\multicolumn{3}{c}{Model Configuration}  &     & \multicolumn{3}{c}{Weight only Quant.} && \multicolumn{3}{c}{{Weight + Activation Quant.}} \\
\cmidrule{1-3} \cmidrule{5-7} \cmidrule{9-11}
\multicolumn{1}{l}{{Model}} & \multicolumn{1}{l}{{Task}} & \multicolumn{1}{l}{Metric} & FP Score &  W OT\%          & Score         & Err           && A OT\%             & Score             & Err              \\ \midrule
BERT-Base                 & MNLI                     & Acc-m                       & 84.44 & 1.6           & 84.80          & -0.36          && 4.5              & 84.22              & 0.22           \\ \midrule 
BERT-Large                & MNLI                     & Acc-m                       & 86.65 & 1.51          & 86.39         & 0.26           && 4                & 85.69              & 0.96             \\
BERT-Large                & STS-B                    & Spearman                    & 90.25 & 1.51          & 90.12         & 0.13           && 2.5              & 89.51              & 0.74              \\
BERT-Large                & SQuAD                    & F1                          & 93.15 & 1.54          & 93.17          & -0.02          && 1.7              & 92.22             & 0.93             \\ \midrule 
RoBERTa-Large             & MNLI                     & Acc-m                       & 90.58 & 1.48          & 90.38         & 0.20            && 4.1              & 89.81              & 0.77              \\
RoBERTa-Large             & STS-B                    & Spearman                    & 92.41 & 1.48          & 92.25        & 0.16           && 4.4              & 91.52              & 0.89             \\
RoBERTa-Large             & SQuAD                    & F1                          & 93.56 & 1.48          & 93.25         & 0.31           && 2.9              & 92.58              & 0.98              \\ \midrule 
DeBERTa-XL                & MNLI                     & Acc-m                       & 91.75 & 1.2           & 91.78        & -0.03          && 4.3              & 91.18              & 0.57   \\   \bottomrule[1pt]          
\end{tabular}
\end{table*}

Table~\ref{tbl:accuracy} reports the resulting task performance under the ``Score'' columns. The ``FP Score'' column reports the task performance of the original FP32/FP16 models, whereas the other two score columns report task performance when we quantize weights alone or weights and activations. The higher the score, the better the model performs. The change in task performance is also listed under the ``Err'' columns. Overall, \OURL is within 1\% of the FP baseline and closer for the most recent DeBERTa. The table also lists the fraction of outlier values in weights (W OT\%) and activations (A OT\%). In the worst case, outliers account for just 2.5\% of the overall values. In some applications, there may be enough on-chip buffering to store the activations as-is. Accordingly, the table also reports task performance when \OURL is applied only to weights. Task performance is, as expected, slightly better and, in some cases, even better than the FP baseline. Such variations in task performance are normal. The outlier fraction is also noticeably lower for the weights, which is also expected as the activations exhibit a much larger range, as corroborated by the use of longer datatypes for activations in prior quantization~work.

\begin{figure}
\centering
\includegraphics[width=0.35\textwidth]{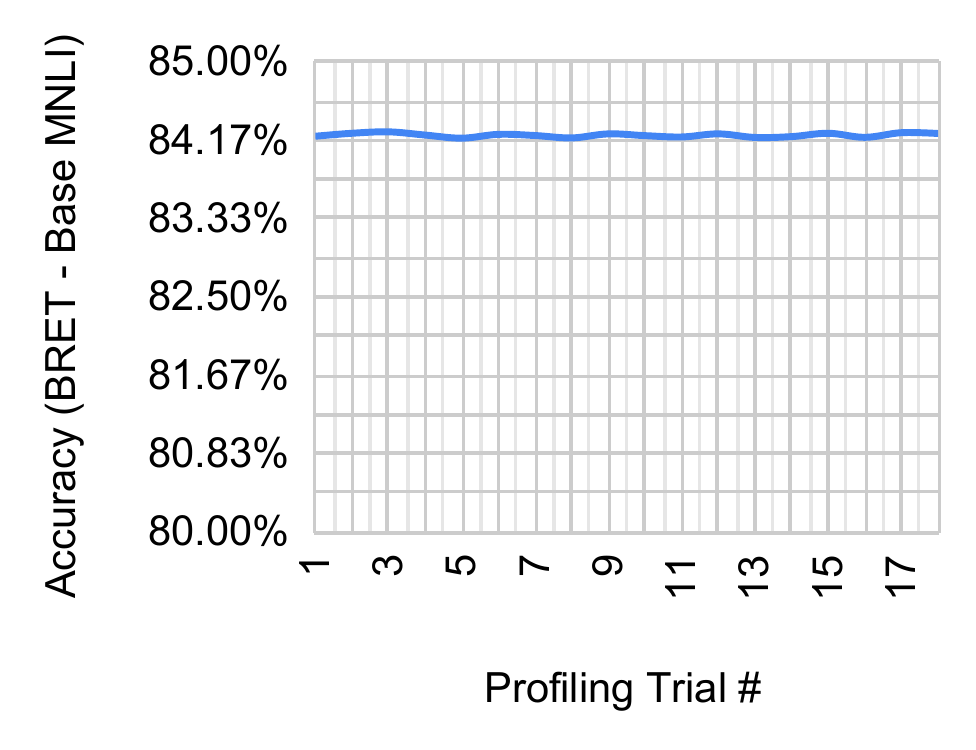}
\caption{Profiling effect on accuracy.}
\label{fig:prof_trial}
\end{figure}

Our experiments show profiling has almost no effect on the accuracy:~Figure~\ref{fig:prof_trial} shows the accuracy after profiling model multiple times with different random samples of the training set and the result of profiling is almost identical each time. The reasons behind this are \textit{I)}~In our method, the bins get exponentially wider when the index ($a^{index}$) gets larger. This means the bins that will contain outliers are wide and capture an extremely wide range of values. \textit{II)}~weights in the model act as a regularizer for the activations and shape the output distribution. Also, transformer-based models have layer normalization and softmax which limits the range of values.

\subsection{Hardware Evaluation Methodology}\label{sec:eval:hw}
As our baseline, we use a {spatial} 2048 FP16 Tensor Cores-based accelerator~\cite{TC1,TC2} and we sweep a range of possible on-chip buffer capacities. We use this baseline to demonstrate that \OURL compression alone can benefit an existing accelerator and to also demonstrate that a design using the \OURL processing units can outperform it. We also compare with GOBO~\cite{GOBO}. The dataflow for all designs is optimized to minimize the number of off-chip transactions.

Cycle counts are measured using a custom cycle-accurate simulator that uses DRAMSIM3~\cite{Dramsim3} to model DRAM transactions for a DDR4-3200 dual-channel main memory. The simulator has been tested with various microbenchmarks and verifies the correctness of the simulated output. Memories are divided into banks to meet the target access time and bandwidth. Area and power for the on-chip memories are modeled using CACTI~\cite{CACTI}.

We use \textit{post-layout} energy and area estimates. All designs are implemented in Verilog. Major arithmetic units are instantiated from Synopsys DesignWare Building Block IP Library~\cite{SynopsysDW}. For synthesis, we use Synopsys' Design Compiler (ver. N-2017.09)~\cite{SynopsysDC} with a 65nm TSMC technology node targeting 1 GHz frequency and for placement and routing, we use Cadence Innovus (ver. 20.13)~\cite{CadenceInnovus}. Power consumption is estimated in Innovus using the post-layout netlist with an activity factor generated by Intel ModelSim.

\subsection{\OURLCORE\ Acceleration}
\label{SEC:HW_accelerator}
The baseline architecture uses FP16 --- an FP32 baseline which is more common today, would have resulted in higher benefits for \OURL. 
Table~\ref{tbl:HW_sum} reports area used by their processing elements and their count for the baseline Tensor Cores, GOBO (refer to section~\ref{sec:related}) and \OURL.
Since the area of each \OURL processing element (PE) is smaller than either those of Tensor Cores or GOBO, \OURL can afford to pack more elements within less area. Moreover, since \OURL stores values using short indexes, it does not need wide memories to supply its units. For example, the \OURL PE is 39\% smaller compared to a tensor-core unit with an equivalent compute-capability (MAC/cycle). This superior processing power per area and energy vs. the baseline is an additional advantage of \OURL.

\begin{table}
\ra{0.9}
\scriptsize
\centering
\caption{Area, Cycle Count and Energy for BERT-Base.}
\begin{tabular}{@{}lccccc@{}} \toprule[1pt]
             & \multicolumn{2}{c}{Compute} && \multicolumn{2}{c}{{BERT-Base/512KB buffer}} \\ 
\cmidrule{2-3} \cmidrule{5-6}     
Architecture & Units         & Area ($mm^2$)         && Cycle Count               & Energy (J)              \\ \midrule
FP16 Tensor Cores  & 2048          & 16.1         && 167M                      & 0.36                    \\
FP16 GOBO         & 2560          & 15.9         && 52M                       & 0.17                    \\
Mokey        & 3072          & 14.8         && 29M                       & 0.09                   \\ \bottomrule[1pt]
\end{tabular}
\label{tbl:HW_sum}
\end{table}

\begin{figure}
\centering
\includegraphics[width=0.45\textwidth]{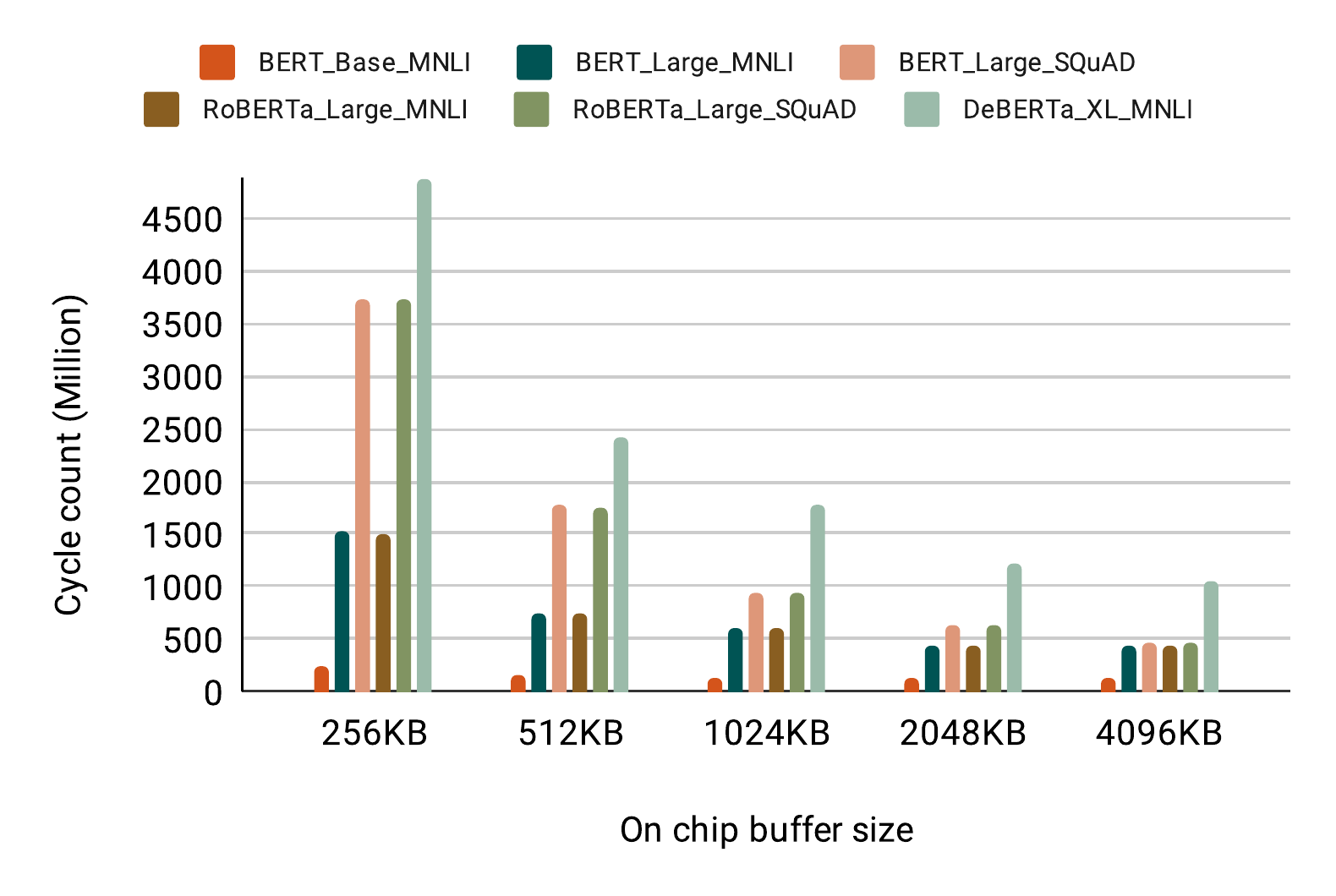}
\caption{Baseline Accelerator: Inference Cycle Counts.}
\label{fig:relative_cycle}
\end{figure}

\noindent\textbf{Comparison with Tensor Cores: }We first compare \OURL with the baseline Tensor Cores accelerator reporting execution time performance and energy efficiency. All measurements are normalized over the corresponding baseline configuration. Tensor Cores and \OURL use the same compute area and the same buffer capacity -- total chip area for \OURL is smaller as it uses narrower buffers and interconnect.
Figure~\ref{fig:relative_cycle} reports the execution time in cycles for the baseline configurations. As expected, using larger on-chip buffers decreases execution time as 1)~it allows for more overlap between off-chip accesses and on-chip compute, and 2)~reduces the number of off-chip transactions due to higher reuse.

\begin{figure*}[t!]
\centering
\begin{minipage}{0.53\textwidth} 
\includegraphics[width=0.95\textwidth]{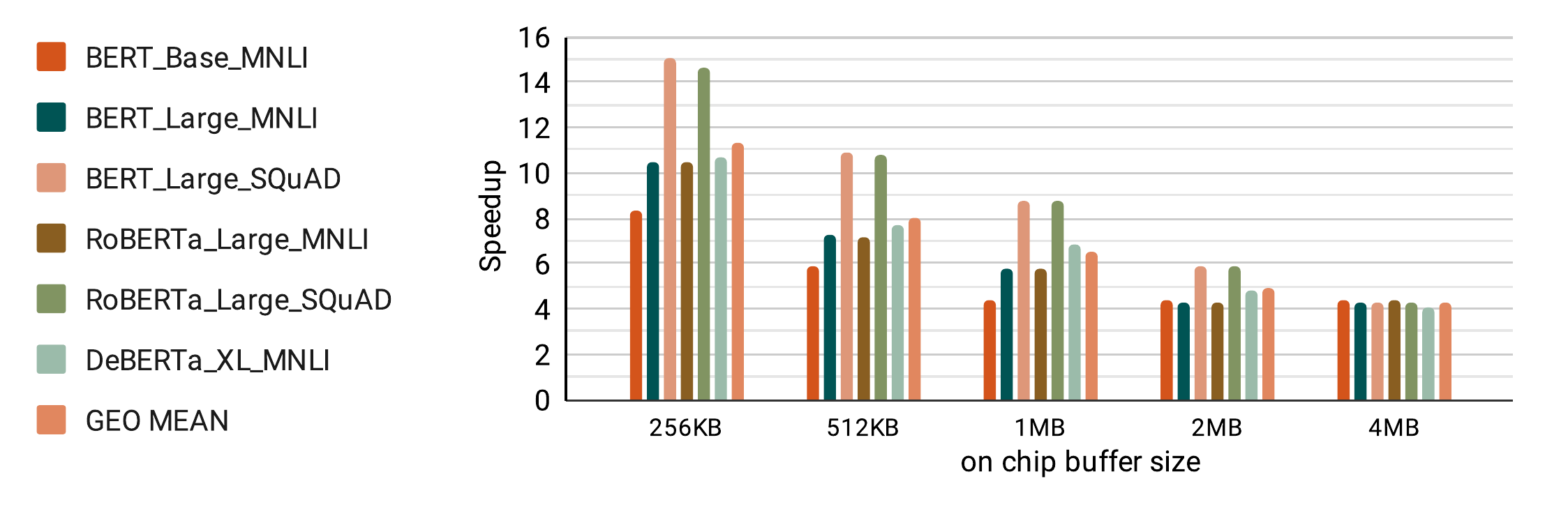}
\caption{\OURL Accelerator Speedup over Tensor Cores.} 
\label{fig:Mokey_comp_perf}
\end{minipage}\hspace{0pt}
\begin{minipage}{0.41\textwidth}
\centering
\includegraphics[width=0.9\textwidth]{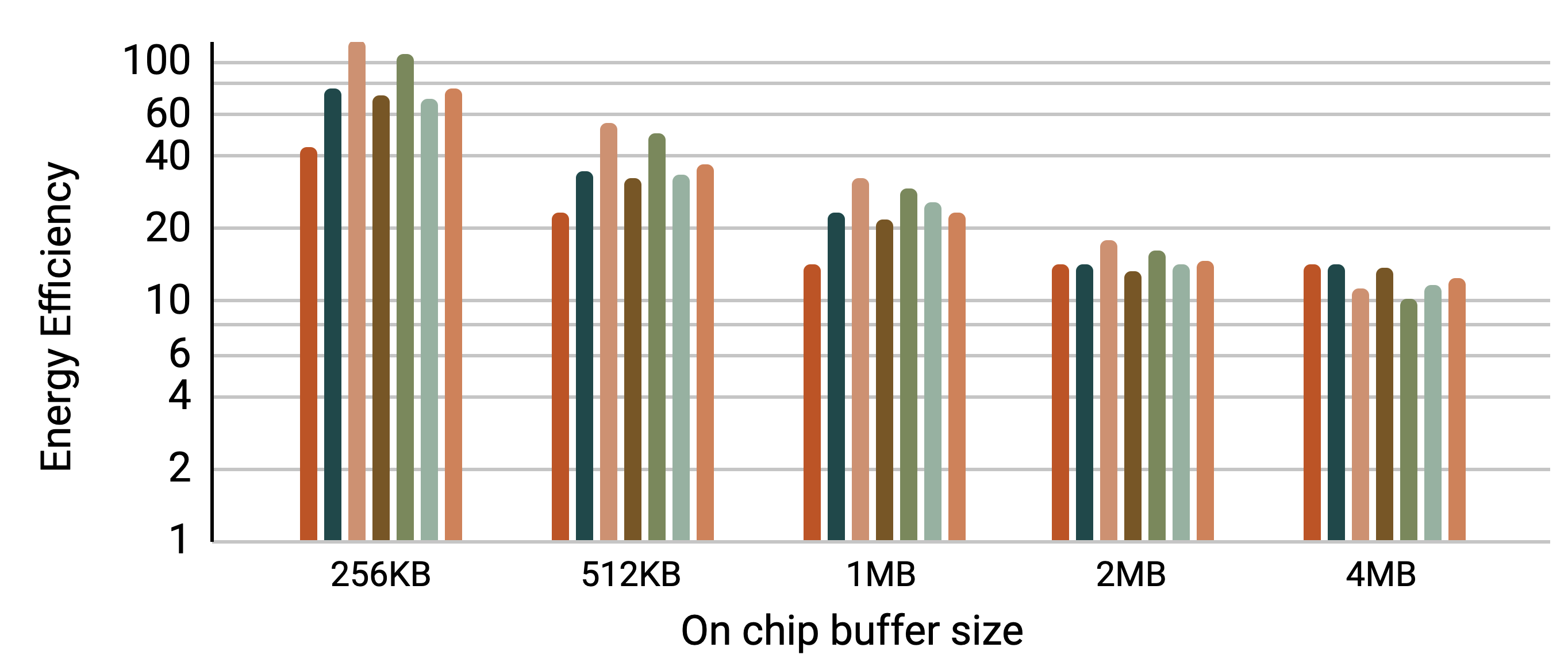}
\caption{\OURL Energy Efficiency over Tensor Cores.}
\label{fig:MokeyHW_EFF}
\end{minipage}\hspace{5pt}
\end{figure*}

Figure~\ref{fig:Mokey_comp_perf} shows that \OURL reduces execution time compared to the baseline. Although \OURL has some overheads for computing outliers and for the post-processing stage, on average it outperforms the baseline by $11\times$ when the buffers are small and by $4.1\times$ faster for the large 4MB buffers.

\begin{figure*}[t!]
\centering
\begin{minipage}{0.53\textwidth} 
\includegraphics[width=0.95\textwidth]{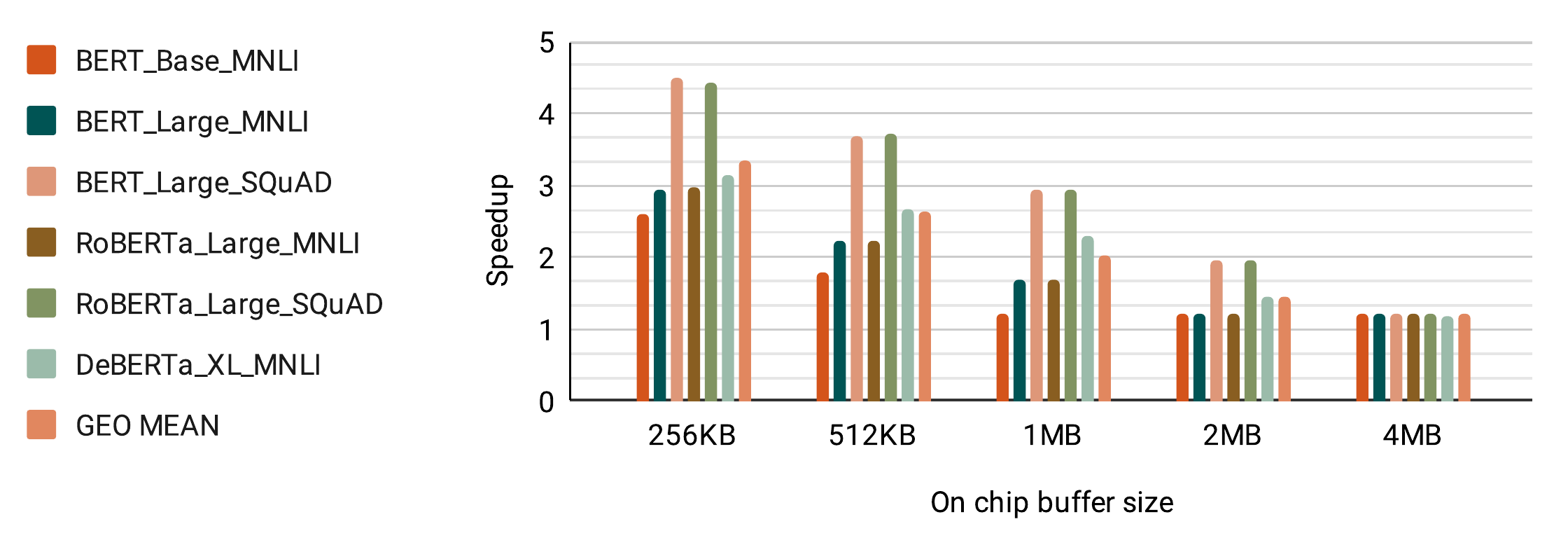}
\caption{\OURL Accelerator Speedup over GOBO.} 
\label{fig:GOBO_PERF}
\end{minipage}\hspace{0pt}
\begin{minipage}{0.41\textwidth}
\centering
\includegraphics[width=0.9\textwidth]{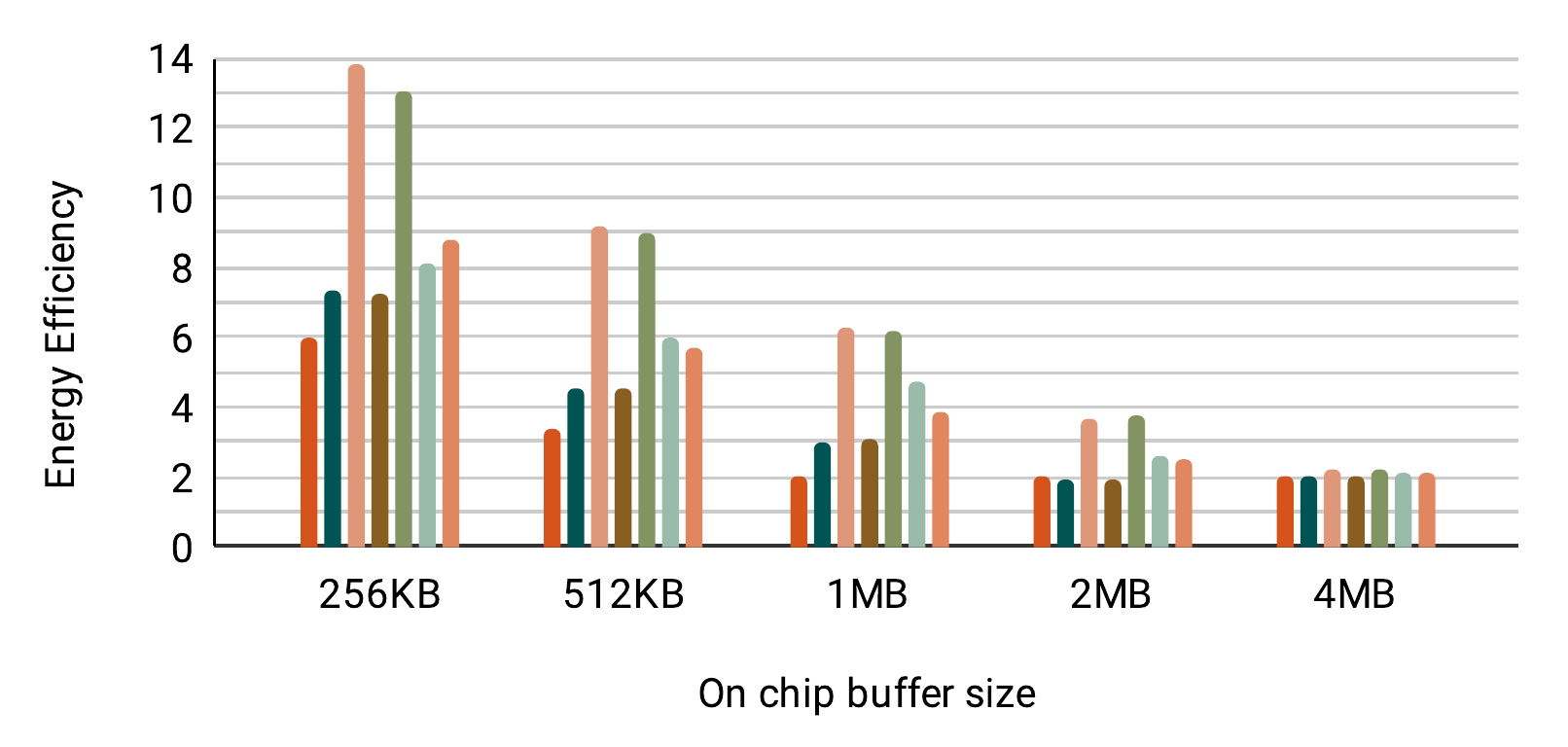}
\caption{\OURL Energy Efficiency over GOBO.}
\label{fig:GOBO_Effic}
\end{minipage}\hspace{5pt}
\end{figure*}

Figure~\ref{fig:MokeyHW_EFF} also reports the resulting relative energy efficiency for the \OURL accelerator. Even for the 4MB on-chip buffer configurations, the \OURL hardware accelerator boosts energy efficiency by $13\times$. For the 256KB configurations \OURL is  $78\times$ more energy efficient.

\begin{table*}
\centering
\caption{{Area, Performance and Energy breakdown for Tensor Cores and \OURL for BERT Large on SQuAD.}}
\ra{0.95}
\begin{tabular}{@{}lcccccccc@{}}\toprule[1pt]
& \multicolumn{2}{c}{256KB on-chip buffer} & \phantom{abc}& \multicolumn{2}{c}{512KB on-chip buffer} &\phantom{abc} & \multicolumn{2}{c}{1MB on-chip buffer} \\ 
\cmidrule{2-3} \cmidrule{5-6} \cmidrule{8-9}
& Tensor Cores                       & \OURL         && Tensor Cores  & \OURL    && Tensor Cores  & \OURL  \\ \midrule
\textbf{Area ($mm^2$)}\\
\phantom{abc}On-chip Buffer Area                            & 13.2     & 4.7        && 16.8     & 8.0        && 24.7     & 14.6         \\
\phantom{abc}Compute Area                                 & 16.1     & 14.8        && 16.1     & 14.8        && 16.1      & 14.8         \\
\phantom{abc}Total Chip Area                             & \textbf{30.7}     & 20.5        && 34.5     & 23.9        && 42.7      & \textbf{30.8}  \\  \midrule  
\textbf{Performance}\\
\phantom{abc}Memory Transfer Cycles                                                & 3690 M   & 226 M       && 1730 M   & 151 M       && 938 M     & 108 M        \\
\phantom{abc}Compute Cycles                                                        & 60 M     & 55 M        && 60 M     & 55 M        && 60 M      & 55 M         \\
\phantom{abc}Total Cycles                                                          & 3734 M   & 249 M       && 1772 M   & 163 M       && 953 M     & 109 M        \\
\phantom{abc}Compute/Memory Avg. Overlap \%                                             & 26.7\%   & 57.7\%      && 29.0\%   & 77.0\%      && 76.5\%    & 98.2\%       \\ \midrule
\textbf{Energy (J)}\\
\phantom{abc}Off-chip Memory Energy                                             & 5.79     & 0.35        && 2.72     & 0.24        && 1.47      & 0.17         \\
\phantom{abc}On-chip Memory Energy                                              & 0.1      & 0.01        && 0.1      & 0.01        && 0.11      & 0.02         \\
\phantom{abc}Compute Energy                                                     & 0.95     & 0.48        && 0.95     & 0.48        && 0.95      & 0.48         \\
\phantom{abc}Total Energy                                                      & 6.84     & 0.84        && 3.77     & 0.73        && 2.53      & 0.67     \\   \bottomrule[1pt]
\end{tabular}
\label{tbl:Breakdown}
\end{table*}

\noindent\textbf{{Advantages of \OURL Hardware Accelerator:} }
{Table}~\ref{tbl:Breakdown} {shows breakdowns of area, performance, and energy for Tensor Cores and \OURL with BERT Large on SQuAD. \textbf{1)~Area:} Since \OURL PEs are 39\% smaller, more can fit in the same compute area. Furthermore, since \OURL computes directly on quantized indexes, it requires on-chip buffers with significantly narrower data \textit{interfaces} and \textit{interconnects} with area and energy benefits. For example, due to the narrower interface, \OURL's 1MB buffers use as much area as the 256KB buffers of Tensor Cores. 
However, in this evaluation, we did \textit{not} use this to \OURL's advantage, e.g., by introducing more units. Instead we used an \textit{iso-compute-area} and \textit{iso-buffer-capacity} constraint and showed that \OURL outperforms Tensor Cores even under this \textit{worse} for \OURL constraint. If we were to account for this advantage of \OURL, combined the $4\times$ reduction in memory area and the reduction of values from 16b to 5b, would translate into a nearly $13\times$ amplification of on-chip memory capacity.  Section}~\ref{sec:componly} shows that the \OURL accelerator is still better than the Tensor Cores even when the latter uses \OURL just for on- and off-chip memory compression. Processing with Tensor Cores requires converting the indexes into the corresponding centroids via lookup tables.

\textbf{2)~Performance:} \OURL's on-chip compression allows more values to be stored on-chip and increases the overlap of compute and memory transactions. For example, with the 256KB on-chip buffer, \OURL increases compute/memory overlap from 27\% to 58\%. Also, since \OURL can fit more weights and activations on-chip it can better reuse values and reduce the overall off-chip traffic. \textbf{3)~Energy:} \OURL compute units are extremely energy efficient. \OURL, despite having $1.5\times$ more compute units, it burns $2\times$ less energy as they perform mostly counting on narrow fixed-point indexes.

\noindent\textbf{Comparison with GOBO: }
Figure~\ref{fig:GOBO_PERF} shows that \OURL is faster than the GOBO accelerator. The figure shows \OURL's execution time normalized over GOBO with the same on-chip buffer size. Whereas GOBO stores activations in FP16, \OURL quantizes them to 4b. This magnifies on-chip buffer capacity, boosting performance for tasks with longer sequence lengths (e.g., SQuAD). With larger on-chip buffers, \OURL remains faster albeit at a lesser degree since the source of performance benefits shifts more to the difference in peak compute capability. Another advantage of \OURL is that its PEs are more efficient as they perform fixed-point operations vs. GOBO's FP16 PEs. This is shown in Figure~\ref{fig:GOBO_Effic} which reports the overall energy efficiency of \OURL over GOBO. \OURL consumes less energy and is $9\times$ more energy efficient compared to GOBO with smaller buffers. The energy efficiency remains high at  $2\times$ even with the 4MB on-chip buffers.

\subsection{Using \OURL for Memory Compression Only}\label{sec:componly}
We consider using \OURL only as compression mechanism (weights and activations) over the Tensor Cores baseline. Larger on-chip buffers store more values on-chip, resulting in 1)~better overlap between computation and memory transactions, and 2)~a reduction in DRAM transactions. 

\begin{figure*}
\centering
\includegraphics[width=0.85\textwidth]{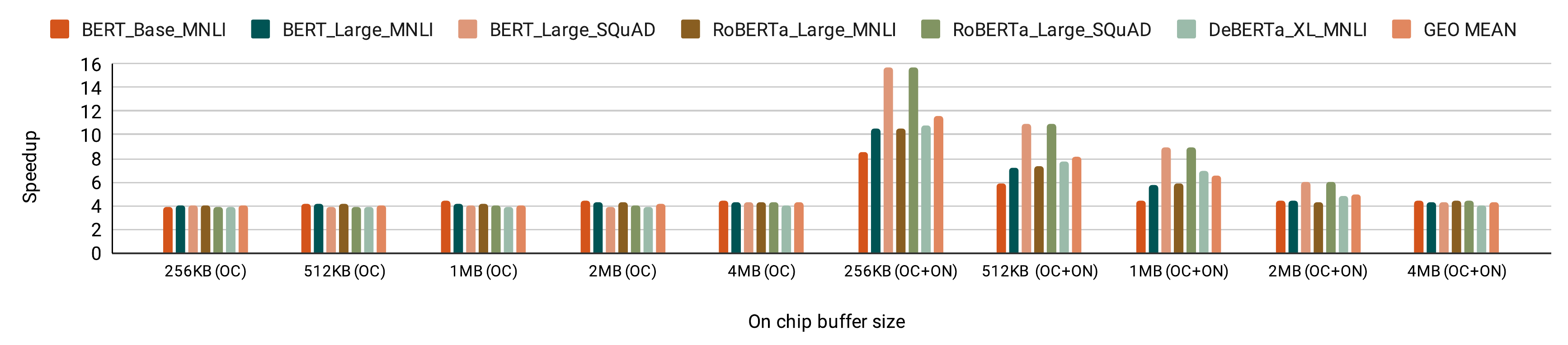}
\caption{Tensor Cores speedup w/ \OURL Memory compression for Off-chip only (OC) and Off- and On-chip (OC+ON) traffic.}
\label{fig:Mem_comp_perf}
\centering
\includegraphics[width=0.85\textwidth]{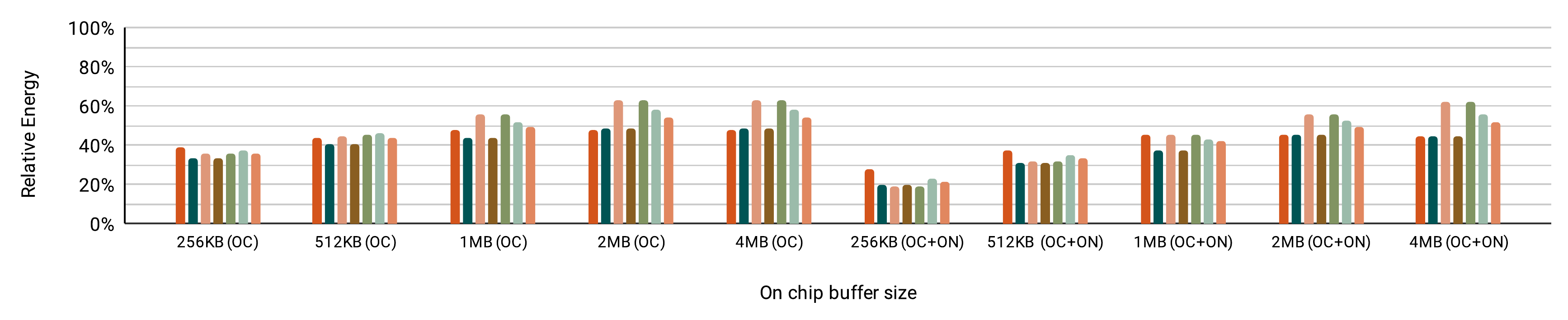}
\caption{Tensor Cores Relative Energy w/ \OURL Memory compression for Off-chip only (OC) and Off- and On-chip (OC+ON).}
\label{fig:Mem_comp_energy}
\end{figure*}

\noindent\textbf{\OURL Off-Chip Compression: } In the first set of experiments, \OURL compresses values off-chip only. When the values are transferred to the chip, they get expanded into FP16 centroids. Figure~\ref{fig:Mem_comp_perf} shows performance normalized over the corresponding baseline configurations that do not use \OURL compression. The average speedup for a small 256KB buffer is about $3.9\times$ and it improves when the on-chip buffer gets larger to $4.3\times$ for a 4MB buffer. These models are mostly memory bound, and with the smaller on-chip buffers, compute units consume the data faster than DRAM transactions resulting in idle time. These configurations most of the time cannot overlap computation with off-chip memory transactions. As the on-chip capacity grows larger, computation and off-chip access overlap becomes increasingly frequent.

In general, off-chip compression reduces DRAM energy consumption by a factor of $4\times$ reducing overall relative energy (Figure~\ref{fig:Mem_comp_energy}). With the 256KB on-chip buffer, on average 82\% of total energy is due to memory transactions. This is less with the larger on-chip buffers as they access DRAM less. With the 4MB on-chip buffer, memory transactions account for 53\% of energy consumption. Regardless, \OURL off-chip compression improves energy efficiency of the 256KB configuration by $11\times$ and by $7.8\times$ for the 4MB configuration. 

\begin{table*}
\ra{0.9}
\centering
\caption{Comparing Various Quantization Methods for BERT-Base on the MNLI task. \textit{INT Comp} indicates if the computation is in fixed-point domain. \textit{Compression Ratio} reports the total memory footprint reduction.}
\begin{tabular}{@{}lccccccc@{}}  \toprule[1pt]
BERT-Base MNLI & Parameters (bit) & Activations (bit) & Accuracy (m) & Error & INT Comp  & Post-Training & Compression Ratio \\ \midrule
FP32 Baseline  & 32      & 32          & 84.44    & -     & \xmark                   & \xmark             & $1\times$                      \\ \midrule
Q8BERT~\cite{intel8b}         & 8       & 8           & 83.75    & 0.69  & \cmark                   & \xmark             & $4.0\times$                    \\
I-BERT~\cite{kim2021ibert}         & 8       & 8           & 84.12    & 0.32  & \cmark                   & \xmark             & $4.0\times$                    \\
Q-BERT~\cite{q-bert}         & 4       & 8           & 83.89    & 0.55  & \xmark                   & \xmark             & $6.9\times$                    \\
GOBO~\cite{GOBO}           & 3       & 32          & 83.76    & 0.68  & \xmark                   & \cmark             & $4.1\times$                    \\
TernaryBERT~\cite{zhang2020ternarybert}    & 2       & 8           & 83.30    & 1.14  & \xmark                   & \xmark             & $10.8\times$                   \\
\OURL          & 4       & \textbf{4}           & 84.22    & \textbf{0.22}  & \cmark                   & \cmark             & $7.9\times$                  \\ \bottomrule[1pt]
\end{tabular}
\label{tbl:Relwork}
\end{table*}

\noindent\textbf{\OURL On-Chip Compression: }
In this application, \OURL stores 5-bit indexes in on-chip buffers and expands the values once they are requested by compute units.
In Figure~\ref{fig:Mem_comp_perf}~\textit{(OC+ON)} set of bars report the corresponding speedup.
Compressing data allows us to store more weights and activations, maximize reuse and significantly reduces the number of DRAM transactions. The smaller the on-chip capacity, the higher the potential benefits from on-chip compression. 
In transformer models, activations grow quadratically with respect to sequence length. BERT-Large and RoBERTa-Large on the SQuAD task used a sequence length of 384 tokens, while for other model/tasks use a sequence length of 128. This explains why these two models on SQuAD task benefit the most from on-chip compression in the smaller buffer sizes. As the on-chip buffer gets larger, eventually, all activations would fit on-chip and the weights need to be loaded only once from off-chip. With \OURL compression, the on-chip buffer size at which this becomes possible is smaller.

Energy reduction with on-chip memory compression exhibits a similar trend as with off-chip as shown in Figure~\ref{fig:Mem_comp_energy}. Energy reduction is significant with the smaller on-chip buffers as \OURL boosts their effective capacity reducing off-chip accesses. With the 256KB buffer \OURL improves energy efficiency by $54\times$ and by $8\times$ with the 4MB buffer.

\section{Related Work}
\label{sec:related}

This section discusses how \OURL compares to state-of-the-art model compression methods for Transformer models.

\noindent\textbf{Integer/Floating-Point Quantization: }
Q8BERT quantized weights and activations to 8-bit fixed-point. It requires fine-tuning to reduce quantization error and FP32 compute cores for some layers such as Softmax~\cite{intel8b}. I-BERT also quantizes weights and activations to 8 bits, but by leveraging approximations for nonlinear functions it uses fixed-point throughout~\cite{kim2021ibert}. Table~\ref{tbl:Relwork} compares Q8BERT and I-BERT with \OURL. Mokey is a post-training method and can be applied on out-of-the-box models without requiring access to datasets or systems that are capable of training these models.  \OURL reduces activation and weight footprint by approximately $2\times$ compared to Q8BERT and I-BERT, has higher accuracy, and performs all computations in fixed-point with most using 3b.

\noindent\textbf{Dictionary-Based Quantization: }
Q-BERT is a group-wise dictionary-based quantization method~\cite{q-bert}. It divides the parameters of each layer into groups (typically 128) each of which it quantizes to a dictionary of 4 to 16 representative values. Q-BERT quantizes activations to 8b. It requires fine-tuning. GOBO is a post-training dictionary-based quantization method that quantizes weights only to 3b or 4b (activations remain in FP16 or FP32)~\cite{GOBO}. GOBO uses an iterative centroid selection method similar to k-means. Table~\ref{tbl:Relwork} compares Q-BERT and GOBO with \OURL. Compared to Q-BERT, \OURL requires no fine-tuning, reduces footprint for activations more, and has better accuracy. Compared to GOBO, \OURL quantizes both weights and activations, achieves higher accuracy, and replaces floating-point arithmetic with narrow fixed-point operations. \OURL quantizes layers based on a Golden Dictionary which is faster than GOBO's iterative process.

\noindent\textbf{Pruning, Distillation and NAS: }
Model pruning, Distillation and Neural Architecture Search (NAS) can reduce model size and are orthogonal to quantization methods such as \OURL. 
TernaryBERT uses a combination of knowledge distillation and quantization to compress BERT~\cite{zhang2020ternarybert}. It quantizes weights to 2b and activations to 8b. TernaryBERT's extreme quantization comes at the expense of a $1.14\%$ drop in accuracy, and the distillation process requires training the model while searching for suitable hyperparameters. 
SpAtten prunes less effective cascade heads and tokens from attention layers~\cite{wang2021spatten}.
HAT is a NAS method which alters model architecture to best fit a target hardware configuration~\cite{wang2020hat}. 

\section{Conclusion}
The bulk of the operations and data transfers in state-of-the-art NLP models {are} due to multiply-accumulate operations (MACs) over floating-point values. 
\OURL introduces a unique quantization that maps the original floating-point value space into a few fixed-point values which are a small subset of a larger fixed-point value space. \OURL chooses this subset so that \textit{all} its values fit under some exponential curve.
It is this property that enables \OURL to replace MACs over floating-point tensors, transforming them into counting of 4b centroid indexes without even having to map to the centroid values themselves --- at the end, a few scaling operations using fixed-point arithmetic map the final result into the target fixed-point value space. Since \OURL adjusts the exponential curve to best fit each layer, it does not restrict the set of possible centroid values.
We expect that future work will expand and improve upon quantization methods such as \OURL. These are methods that carefully select not only the target value space, but also a subset within it to maximize cost reduction opportunities.

\section*{Acknowledgement}
This work was supported by the NSERC COHESA Strategic Research Network and an NSERC Discovery Grant. The University of Toronto maintains all rights to the technologies described.

\bibliographystyle{ieeetr}
\bibliography{bibtex/bib/IEEEabrv,bibtex/bib/IEEEexample.bib}

\end{document}